\documentclass{article} % For LaTeX2e
\usepackage{iclr2025_conference,times}

\usepackage{amsmath,amsfonts,bm}

\def\eqref#1{equation~\ref{#1}}
\def\1{\bm{1}}

\DeclareMathAlphabet{\mathsfit}{\encodingdefault}{\sfdefault}{m}{sl}
\SetMathAlphabet{\mathsfit}{bold}{\encodingdefault}{\sfdefault}{bx}{n}

\usepackage[utf8]{inputenc}
\usepackage[T1]{fontenc}
\usepackage{hyperref}
\usepackage{url}
\usepackage{booktabs}
\usepackage{amsfonts}
\usepackage{nicefrac}
\usepackage{microtype}
\usepackage{xcolor}
\usepackage{cleveref}
\usepackage{graphicx}
\usepackage{multirow}
\usepackage{listings}
\usepackage{wrapfig}
\usepackage[normalem]{ulem}
\usepackage{pgffor}

\title{\ours{}: Open-endedness via\\Models of human Notions of Interestingness\\with Environments Programmed in Code}

\author{%
Maxence Faldor\thanks{co-authors}\\
Department of Computing\\
Imperial College London\\
London, United Kingdom\\
\texttt{m.faldor22@imperial.ac.uk} \\
\And
Jenny Zhang\footnotemark[1]\\
Department of Computer Science\\
University of British Columbia\\
Vector Institute\\
\texttt{jennyzzt@cs.ubc.ca}
\And
Antoine Cully\thanks{co-senior authors}\\
Department of Computing\\
Imperial College London\\
London, United Kingdom\\
\texttt{a.cully@imperial.ac.uk}\\
\And
Jeff Clune\footnotemark[2]\\
Department of Computer Science\\
University of British Columbia\\
Vector Institute\\
Canada CIFAR AI Chair\\
\texttt{jeff.clune@ubc.ca}\\
}

\newif\ifcomment
\iclrfinalcopy % Uncomment for camera-ready version, but NOT for submission.
\begin{document}

% \commenttrue % Set \commentfalse to togle inline coments off
\commentfalse

% Algorithms
\newcommand{\ours}{OMNI-EPIC}
\newcommand{\website}{\url{https://dub.sh/omniepic}}
\newcommand{\blue}[1]{\textcolor{blue}{#1}}

\maketitle

\begin{abstract}
Open-ended and AI-generating algorithms aim to continuously \emph{generate} and \emph{solve} increasingly complex tasks indefinitely, offering a promising path toward more general intelligence. To accomplish this grand vision, learning must occur within a vast array of potential tasks. Existing approaches to automatically generating environments are constrained within manually predefined, often narrow distributions of environments, limiting their ability to create \emph{any} learning environment. To address this limitation, we introduce a novel framework, \ours{}, that augments previous work in Open-endedness via Models of human Notions of Interestingness (OMNI) with Environments Programmed in Code (EPIC). \ours{} leverages foundation models to autonomously generate code specifying the next learnable (i.e., not too easy or difficult for the agent’s current skill set) and interesting (e.g., worthwhile and novel) tasks. \ours{} generates both environments (e.g., an obstacle course) and reward functions (e.g., progress through the obstacle course quickly without touching red objects), enabling it, in principle, to create any simulatable learning task. We showcase the explosive creativity of \ours{}, which continuously innovates to suggest new, interesting learning challenges. We also highlight how \ours{} can adapt to reinforcement learning agents’ learning progress, generating tasks that are of suitable difficulty. Overall, \ours{} has the potential to endlessly create learnable and interesting environments, further propelling the development of self-improving AI systems and AI-Generating Algorithms. Project website with videos: \website{}.
\end{abstract}

\vspace{-4pt}
\section{Introduction}
\label{sec:introduction}
\vspace{-4pt}
% Motivation
% AI makes progress
In recent years, the field of artificial intelligence (AI) has witnessed groundbreaking achievements, particularly with advancements in reinforcement learning (RL)~\citep{openai_SolvingRubikCube_2019, silver_MasteringGameGo_2016, colas2019curious} and foundation models (FMs)~\citep{brown_LanguageModelsAre_2020,radford2019language}.
% But AI is narrow
Yet, the most ambitious goal of AI --- building generalist agents capable of autonomously perceiving, reasoning, deciding, and acting within complex environments --- remains a formidable challenge. The creation of general-purpose agents promises immense societal benefits, provided we can address the critical safety and existential risks involved~\citep{bengio_ManagingAIRisks_2023}, including those unique to open-ended and AI-generating algorithms (AI-GAs)~\citep{clune_AIGAsAIgeneratingAlgorithms_2020, ecoffet_OpenQuestionsCreating_2020}.
% Bottleneck has shifted
While researchers have made substantial progress in developing powerful learning architectures~\citep{hafner_MasteringDiverseDomains_2024, jaegle_PerceiverIOGeneral_2022, vaswani_AttentionAllYou_2017}, these algorithms have been applied to a relatively narrow range of tasks due to the limited availability of diverse datasets. Compounding this issue, current models require extensive amounts of data and fine-tuning to be trained. Consequently, over the past decade, the bottleneck in advancing AI has shifted from improving learning algorithms to acquiring the necessary data to train them~\citep{jiang_GeneralIntelligenceRequires_2022}. In other words, the main challenge has become: how to effectively acquire large amounts of data for diverse tasks.

% Different approach: Open-endedness
A fundamentally different approach involves developing open-ended algorithms that continuously generate and solve new challenges endlessly~\citep{stanley2017open}. The objective of open-ended algorithms is to ignite an explosion of creativity and complexity in a computer, akin to the processes observed in biological evolution and human culture, including science and technology. Replicating open-ended evolution \emph{in silico} extends beyond understanding biological life or developing engaging simulations. It delves into fundamental questions about the nature of creativity, the potential for machines to exhibit life-like characteristics, and the emergence of general intelligence. Given that biological evolution is the only known process that has successfully created general intelligence, its principles hold invaluable insights for advancing the field of AI, and creating AI-GAs~\citep{clune_AIGAsAIgeneratingAlgorithms_2020}.

% Darwin Completeness
For open-ended algorithms and AI-GAs to succeed, they must operate within a \emph{vast} task space capable of generating an infinite array of potential challenges. Previous open-ended algorithms~\citep{sudhakaran_MarioGPTOpenEndedText2Level_2023,wang2023voyager,wang_PairedOpenEndedTrailblazer_2019,wang_EnhancedPOETOpenended_2020,zhang_OMNIOpenendednessModels_2023} have been applied to limited domains with specific types of worlds (e.g., obstacle courses) and confined to predefined parameterizations (e.g., obstacle size and type), restricting their potential to exhibit true open-endedness. Achieving \emph{Darwin Completeness} --- the potential to generate \emph{any possible learning environment} --- is essential for realizing the full potential of AI-GAs~\citep{clune_AIGAsAIgeneratingAlgorithms_2020}.
However, Darwin Completeness presents a significant challenge: the ability to generate infinitely many environments makes the search potentially intractable, or even impossible. Exploring this immense environment search space may involve endlessly generating trivial, redundant, or overly complex tasks that do not effectively contribute to the agent's learning progress. The main challenges are to ensure that generated environments are novel, interesting, solvable, and appropriately matched to the current capabilities of the learning agents. Biological evolution required an unfathomable amount of computation over billions of years to produce intelligence. Therefore, one key scientific challenge is to determine how we can optimize the process of generating and solving interesting tasks within a Darwin Complete environment search space, so as to create an AI-GA given the computational capabilities we expect to have in the future.

% Methods
Previous work in Open-endedness via Models of human Notions of Interestingness (OMNI)~\citep{zhang_OMNIOpenendednessModels_2023} leverages FMs to improve open-ended learning by focusing on tasks that are both learnable and interesting. However, OMNI, like all prior open-ended works~\citep{sudhakaran_MarioGPTOpenEndedText2Level_2023,wang_PairedOpenEndedTrailblazer_2019,wang_EnhancedPOETOpenended_2020,zhang_OMNIOpenendednessModels_2023,wang2023voyager}, was confined to generating tasks within a narrow environment search space, inhibiting the generation of \emph{any} possible learning environment. This paper introduces a novel framework, \ours{}, that augments OMNI with Environments Programmed in Code (EPIC). \ours{} utilizes FMs to choose the next interesting and learnable task and subsequently generate environment code to enable the agent to learn how to solve that task. Our approach generates not only the simulated world but also the reward and termination functions, allowing it, in principle, to create any simulatable task. We take advantage of pre-existing simulators and \ours{} writes code to create tasks within it. For example, if the task is to kick a ball to hit a moving target, \ours{} would generate the environment code to simulate the physics, the agent, the ball, and the moving target, rewarding the agent when the ball hits the target. Conversely, for a task involving maneuvering the ball around a moving target, the simulated world remains the same, but the reward function could differ, penalizing the agent for any contact with the target. A model of interestingness (MoI) is employed both when generating the next task and checking if any newly proposed task is interestingly new compared to similar ones in the archive. Finally, we introduce a success detector that can automatically determine whether the agent has successfully completed any proposed task.

% Connect back to AI-GA
Our vision is for this algorithm to generate any code, including installing and modifying any existing simulator, or even writing the code for a new simulator. Given that the programming language used here (Python) is Turing complete, \ours{} could potentially create any computable environment (e.g., logic and math problems to quests in virtual worlds, such as building a computer in Minecraft). As a first step toward this ambitious goal, in this work, we constrain our method to write code for one simulator, namely PyBullet~\citep{coumans2016pybullet}. By continuously generating learnable and interesting environments, \ours{} advances the development of self-improving AI systems, bringing us closer to achieving Darwin Completeness and realizing AI-GAs.

\vspace{-4pt}
\section{Related Work}
\label{sec:related-work}
\vspace{-4pt}

\textbf{Unsupervised Environment Design.}
Unsupervised environment design has garnered increasing interest in RL. Several works have explored the development of auto-curricula to perpetually generate new training environments for agents~\citep{dennis_EmergentComplexityZeroshot_2020, jiang_PrioritizedLevelReplay_2021, parker-holder_EvolvingCurriculaRegretBased_2022, samvelyan_MAESTROOpenEndedEnvironment_2022, wang_PairedOpenEndedTrailblazer_2019, wang_EnhancedPOETOpenended_2020}. However, a significant limitation of these methods is their reliance on predefined or manually curated distributions of tasks or environment parameters~\citep{dennis_EmergentComplexityZeroshot_2020, heess_EmergenceLocomotionBehaviours_2017, jiang_PrioritizedLevelReplay_2021, parker-holder_EvolvingCurriculaRegretBased_2022, samvelyan_MAESTROOpenEndedEnvironment_2022, wang_PairedOpenEndedTrailblazer_2019, wang_EnhancedPOETOpenended_2020}, inhibiting the generation of any possible learning environment.
Regret-based approaches~\citep{jiang_PrioritizedLevelReplay_2021, parker-holder_EvolvingCurriculaRegretBased_2022, samvelyan_MAESTROOpenEndedEnvironment_2022} prioritize tasks with high regret, measured by the difference between the highest known return and the mean return across simulations. Alternative methods calculate learning progress by measuring the difference in the agent's task success rates across training steps~\citep{kanitscheider2021multi}. By focusing on tasks with high learning progress, these approaches aim to guide the agent's learning towards the most promising areas of the task space~\citep{oudeyer2007intrinsic, oudeyer2007intrinsicm, baranes2013active}.
However, a critical challenge remains in distinguishing which environments are genuinely interesting~\citep{jiang_GeneralIntelligenceRequires_2022, zhang_OMNIOpenendednessModels_2023, colas2022autotelic}. In the vast space of any task describable in natural language, there may be countless learnable but not meaningful environments (e.g., kicking a ball into a goal at slightly different positions). Inspired by \citet{zhang_OMNIOpenendednessModels_2023}, \ours{} uses human notions of interestingness distilled into FMs to generate environments that are not only learnable, but also interesting.

\textbf{Foundation Models for Environment Design.}
Recent advancements in FMs have showcased their remarkable ability to capture extensive knowledge across diverse subjects by training on vast text corpora~\citep{bommasani2021opportunities}. Consequently, this has sparked interest in applying FMs to environment design. \citet{ma_EurekaHumanLevelReward_2023} utilize FMs to generate code for reward functions while \citet{wang_GenSimGeneratingRobotic_2023} employ FMs to generate simulation environments and expert demonstrations. However, these methods do not build upon the agent's previous performance on different tasks, and hence lack an auto-curriculum that can provide an endless stream of environments and tasks. In procedural content generation~\citep{shaker2016procedural, juliani2019obstacle, justesen2018illuminating}, \citet{sudhakaran_MarioGPTOpenEndedText2Level_2023} and \citet{todd_LevelGenerationLarge_2023} fine-tune FMs to create domain-specific levels. \citet{bruce_GenieGenerativeInteractive_2024} propose training a world model to generate game environments. These approaches focus on level creation but do not train agents or adapt the difficulty based on the agent's performance or learning progress. \citet{zala_EnvGenGeneratingAdapting_2024} generate environments as a curriculum to learn a fixed set of tasks but is limited in its ability to generate truly open-ended environments and adapt to the agent's evolving capabilities. \citet{wang_RoboGenUnleashingInfinite_2023} rely on a predefined set of objects and is limited in reflecting real-world dynamics in its simulation.
Despite significant progress in applying FMs to environment design, opportunities remain for further exploration, such as integrating the generative capabilities of FMs with adaptive auto-curricula. \ours{} addresses this challenge, aiming to unlock the potential for truly open-ended and effective learning environments.

\textbf{Foundation Models in Open-Endedness.}
The field of open-endedness seeks to create algorithmic systems that produce never-ending innovation~\citep{stanley2017open}. There has been increasing interest in leveraging FMs to generate intelligent variations for code or text in evolutionary algorithms. However, these approaches are often confined to a fixed archive with hand-crafted characteristics~\citep{bradley2023quality,ding2023quality, lehman_EvolutionLargeModels_2024,lim2024large}. The core objective of open-endedness algorithms is to generate and solve an endless stream of tasks. One way to do so is to keep an ever-expanding archive of tasks or solutions. \citet{zhang_OMNIOpenendednessModels_2023} and \citet{wang2023voyager} have adopted FMs as a mechanism for auto-curricula, proposing both learnable and interesting tasks for agent training. By restricting the agent's interactions to a predefined range of environmental conditions, these methods may hinder the development of truly adaptable and versatile agents capable of handling the complexities of real-world scenarios. \ours{} addresses this limitation by leveraging FMs to generate not only tasks but also the simulated worlds and reward functions, potentially exposing agents to a wider range of challenges and learning opportunities.

\vspace{-4pt}
\section{Method}
\label{sec:method}
\vspace{-4pt}
\begin{figure}[ht]
    \centering
    \includegraphics[width=\textwidth]{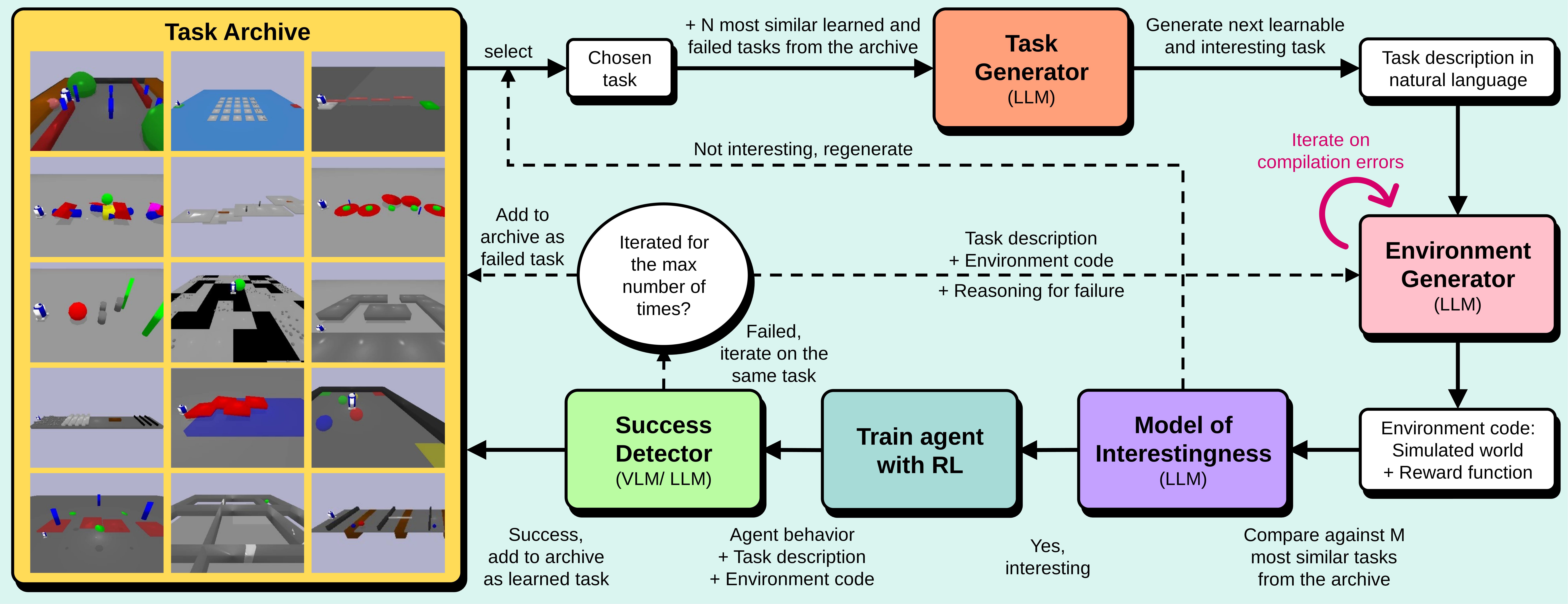}
    \vspace{-20pt}
    \caption{\textbf{\ours{} overview.} \ours{} continuously generates and solves new, interesting tasks in simulation. Our approach maintains a task archive of learned and failed tasks.}
    \label{fig:architecture}
    \vspace{-5pt}
\end{figure}

\ours{} leverages FMs, including large language models (LLMs) and vision-language models (VLMs), to autonomously create learnable and interesting tasks for open-ended learning (\Cref{fig:architecture}). \ours{} maintains a growing task archive (\Cref{subsec:task-archive}) that catalogs successfully learned and completed tasks, as well as unsuccessfully attempted ones. The task generator (\Cref{subsec:task generator}) uses information from the archive about what has been learned and what has not, proposing the next \emph{interestingly new} task, described in natural language, for the agent to attempt.
Because the model has distilled a sense of what is interesting from training on internet data, it has a MoI that emulates the human capacity for making nuanced judgments of interestingness in open-ended learning~\citep{zhang_OMNIOpenendednessModels_2023}. The task generator utilizes this MoI when generating tasks.
These tasks are then translated into environment code by an environment generator (\Cref{subsec:environment generator}), specifying the simulated world and functions required for RL. The newly generated task and its environment code are assessed by a second, post-generation MoI (\Cref{subsec:model-of-interestingness}), to ensure the task is indeed interesting given what has come before (it compares the new task to the most similar tasks already in the archive with retrieval-augmented generation~\citep{lewis2020retrieval}. Tasks deemed interesting are then used to train an RL agent (\Cref{subsec:train-agent-rl}). If deemed uninteresting, the task is discarded, and a new task is generated. After training, a success detector (\Cref{subsec:success-detector}) assesses whether the agent has successfully completed the task. Successfully completed tasks are added to the archive. Failed tasks are iterated upon a maximum number of times and added to the archive as failed tasks if the RL agents are not able to solve them. Then, the cycle of generating the next task restarts. Each component of \ours{} is explained in more detail below, and hyperparameters are shown in \Cref{tab:hyperparameters-ours}. \ours{}'s iterative process ensures continuous generation and learning of new interesting tasks, forming a potentially never-ending growing collection of environments and learned agents.

\vspace{-5pt}
\subsection{Task Archive}
\label{subsec:task-archive}
\vspace{-4pt}

\ours{} maintains a continuously expanding archive of tasks, including successfully learned ones and those attempted but failed. Successful tasks serve as stepping stones for creating more complex yet learnable ones, while failed tasks provide insights into generating new tasks within the agent's current capabilities. \ours{} uses past experiences to generate novel and diverse challenges, continuously pushing the boundaries of what's already learned. The task archive is initialized with a few task description seeds in natural language (\Cref{appendix:task-desc-seeds}). This archive is unbounded and can grow indefinitely as new tasks are generated and learned. Given the generality, interpretability, and universality of natural language and programs~\citep{hopcroft2001introduction}, each task is represented by its natural language description and the corresponding environment, represented as executable code.

\vspace{-5pt}
\subsection{Task Generator}
\label{subsec:task generator}
\vspace{-4pt}

Open-ended algorithms require focusing on tasks that are both learnable (i.e., not too difficult or too easy for the agent to learn) and interesting (i.e., worthwhile and sufficiently novel). Previous attempts have resulted in pathologies when optimizing against definitions and quantifications of interestingness~\citep{zhang_OMNIOpenendednessModels_2023}. Inspired by \citet{zhang_OMNIOpenendednessModels_2023}, we harness FMs to model ineffable human notions of interestingness, gleaned from large text corpora of human-generated data. Here, the task generator is an LLM, which proposes novel task descriptions in natural language that are distinct from those already discovered while remaining learnable (full prompt in \Cref{appendix:task-gen-prompts}).

To ensure the task generator remains open-ended and continuously suggests new and diverse tasks, it uses the content of the task archive as context. Given the limited context length of current LLMs, we retrieve a predefined number of tasks that are most similar to a randomly selected task from the archive (\Cref{appendix:selecting-similar-tasks}). These retrieved tasks include both those that were successfully completed and those that were attempted but failed. These similar tasks serve as examples and are input into the LLM, which then generates the next learnable and interesting task. We opt for the most similar tasks rather than the most different ones to use previous tasks as stepping stones. As LLMs and FMs improve, we expect that a larger portion of the task archive, or potentially the entire archive, could be used as context. However, partial knowledge of stepping stones might be advantageous for creativity and diversity, much as human scientists and artists benefit from not being aware of everything that has come before. The task generator outputs a natural language description of the next task, crafted to be both achievable and interesting for the agent. This description serves as the basis for the subsequent environment generation step, where the natural language task descriptions are translated into executable code to create learning environments.

\vspace{-5pt}
\subsection{Environment Generator}
\label{subsec:environment generator}
\vspace{-4pt}

The environment generator, powered by an LLM, translates a given natural language task description into executable (here, Python) code defining the learning environment. \Cref{appendix:env-gen-prompts} shows the full prompt. This code includes specifications for creating the simulated world and functions needed for RL~\citep{sutton_ReinforcementLearningIntroduction_2018} based on the standard API Gymnasium~\citep{towers_gymnasium_2023}: \texttt{reset}, \texttt{step}, \texttt{reward}, and \texttt{terminated}. The \texttt{reset} function resets the environment to an initial state, including setting up the initial positions and orientations of the agent and objects. The \texttt{step} function updates the environment according to the simulated physics, the agent's action, and any other dynamic behaviors (e.g., moving platforms or activated doors). The \texttt{reward} function returns a scalar number, whose cumulative maximization defines the task. The \texttt{terminated} function indicates whether the agent has reached a terminal state.

For example, if the task is to ``cross a bridge with moving segments'', a bridge should be created in the simulated world. The \texttt{reset} function should initialize the agent at the start of the bridge and each segment's position. The \texttt{step} function should update the environment based on the agent's actions and each segment's movement. The \texttt{reward} function should reward the agent's progress across the bridge, and the \texttt{termination} function should indicate when the agent falls off the bridge.

If compilation errors occur when generating the environment code, the errors (with the traceback) are fed back into the environment generator, which then modifies and improves the environment code. \Cref{appendix:env-gen-reflect-prompts} shows the full prompt. This loop is limited to a maximum of five iterations per task. If the code still fails to compile after these attempts, the uncompiled code is discarded, and the task generator proposes a new task, potentially one that is less complex or differently structured.

\vspace{-5pt}
\subsection{Post-Generation Model of Interestingness}
\label{subsec:model-of-interestingness}
\vspace{-4pt}

While ideally all tasks proposed by the task generator should be interesting (owing to its MoI), the limited context length of current FMs prevents the task generator from considering the entire task archive at once. Furthermore, it is often easier for an FM to evaluate whether a solution is good rather than generate a good solution~\citep{bradley2023quality}.
\ours{} draws inspiration from the dynamics of human culture. For example, researchers often study a specific subset of previous works to inspire new ideas. Once a new idea is produced, it is crucial to check the literature to determine whether the contribution is truly novel. If the idea is deemed interestingly new, it is published, adding to the ever-growing archive of human knowledge. This process creates a growing set of stepping stones to leap off of, which is a key ingredient of open-endedness~\citep{stanley2015greatness}. \ours{} captures these important dynamics of open-ended algorithms, considering a small batch of related stepping stones when creating a new task (\Cref{subsec:task generator}) and then verifying its novelty against the task archive.
Given a newly generated task and its corresponding environment code, we compare it against a predefined number of the most similar tasks from the archive (\Cref{appendix:selecting-similar-tasks}). The post-generation MoI then evaluates if the new task is interesting (e.g., novel, surprising, diverse, worthwhile) (full prompt in \Cref{appendix:model-of-int-prompts}). If the task is deemed interesting, we proceed to train an RL agent on it. If not, the task is discarded, and a new one is generated.

\vspace{-4pt}
\subsection{Training Agents with Reinforcement Learning}
\label{subsec:train-agent-rl}
\vspace{-4pt}
A key objective of open-ended learning is to enable agents to master an ever-expanding set of tasks. To achieve this, \ours{} generates a diverse array of tasks along with their corresponding environments programmed in code. Agents are then trained in these environments using RL to solve the generated tasks. In this work, we utilize the PyBullet physics simulator~\citep{coumans2016pybullet}. While any RL algorithm could be used, we employ DreamerV3~\citep{hafner_MasteringDiverseDomains_2024}. \Cref{tab:hyperparameters-dreamer} details the hyperparameters and compute resources used.
Agents receive proprioceptive (joints positions and velocities) and visual information (images of size $64 \times 64 \times 3$). While \ours{} can be applied to any robot (\Cref{appendix:results-ant}), due to computational limitations, we demonstrate results using an R2D2 robot with a discrete action space of six actions: do nothing, go forward, go backward, rotate clockwise, rotate counterclockwise, and jump. R2D2's simple action space allows the agent to learn tasks more efficiently, requiring fewer time steps than agents with complex action spaces.

Following \citet{wang_EnhancedPOETOpenended_2020}, we train one specialist RL agent for each task. For tasks used to initialize the archive, the RL agents are trained from scratch. For newly generated tasks, the agent continue training from an existing policy previously trained on tasks in the archive. The trained policy is selected from a successfully completed task with the closest embedding to the new task (\Cref{appendix:selecting-similar-tasks}). This approach allows the agents to build upon their existing knowledge and adapt more efficiently to challenges presented by the new tasks. Additionally, color variation in the environments is a form of domain randomization, as an RL agent trained on an environment with specific colors may not perform well in an identical environment with different colors~\citep{tobin2017domain}.

% In this work, we have chosen to use the PyBullet simulator for several reasons. First, PyBullet is a widely adopted open-source simulator. Second, PyBullet offers a convenient and flexible interface for dynamically adding and removing objects from the simulation. Third, PyBullet has been a well-established simulator for an extended period, resulting in a wealth of publicly available code examples and resources. This suggests that FMs are likely to have encountered a substantial volume of PyBullet code during training, making them more adept at generating accurate and functional PyBullet code compared to new or less popular simulators.

\vspace{-4pt}
\subsection{Success Detector}
\label{subsec:success-detector}
\vspace{-4pt}

In this infinite task space of any task describable in natural language, an essential ingredient to training agents to learn any generated task is a universal reward function, which can evaluate if \emph{any} proposed task has been completed or not. We employ a success detector, instantiated as an LLM or VLM, that assesses whether the agent has successfully completed the given task. Since our preliminary testing found that current VLMs are not yet accurate enough to be used as success detectors (\Cref{appendix:vlm-succdet}), we use code generated by LLMs for this purpose instead.

When using an LLM as the success detector, we ask the environment generator to generate an additional success-checking function \texttt{get\_success}, alongside the environment code, that checks if the agent has successfully completed the task. The success-checking function serves a different purpose from the reward function. The reward function is designed to enable the RL agent to learn efficiently, which often results in a more complex function than the success-checking function. The complexity of the reward function arises from the need to shape rewards to facilitate optimal learning~\citep{krakovna2020specification}. Meanwhile, the success-checking function aims to evaluate whether the agent has accomplished the task. This function is less susceptible to reward hacking, as it does not affect how the agent learn.
For example, consider the task of ``run forward''. The success-checking function would evaluate whether the agent's x-axis position has exceeded a certain threshold. In contrast, the reward function should encourage efficient learning by rewarding the agent's x-axis velocity and promoting cyclic joint movements that resemble a natural running motion.

If a task is deemed to have been successfully learned by the trained agent, it is added to the archive. If not, the task is returned to the environment generator for modifications to aid in the agent's learning (e.g., changing the reward function or making the physical world less complex). This feedback loop is repeated for a max number of attempts. If the task remains unlearned after these attempts, it is added to the archive as a failed task and a new task is generated.

\vspace{-4pt}
\section{Long Run with Simulated Learning}
\label{sec:long-run}
\vspace{-4pt}

\begin{figure}[ht]
    \centering
    \includegraphics[width=\textwidth]{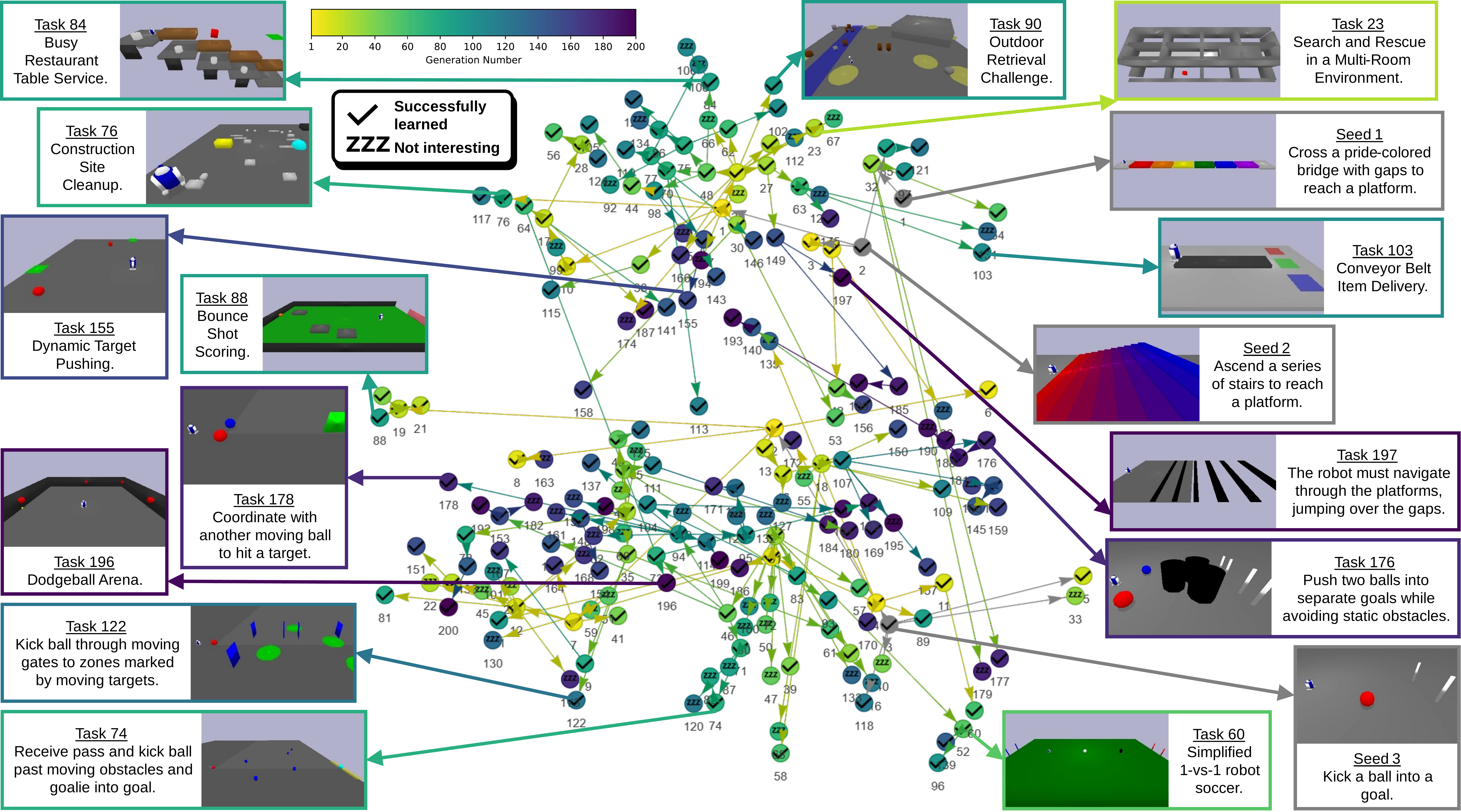}
    \vspace{-20pt}
    \caption{\textbf{Long Run with Simulated Learning.} \ours{} generates a diverse array of tasks, ranging from wildly different objectives to interesting variations of similar overarching tasks. The node color reflects the generation number of the task. A check mark in the node means that the task was successfully learned. A ZZZ symbol means that the task was deemed uninteresting and discarded. The node connections illustrate which tasks were conditioned on when asking an FM to generate a similar yet new and interesting task. Grey nodes show task description seeds that initialized the run.}
    \label{fig:results-long-run}
    \vspace{-10pt}
\end{figure}

To illustrate the creative explosion of generated tasks, we run \ours{} without training RL agents, assuming all generated tasks can be successfully completed. This allows us to showcase a larger number of generated tasks, as training RL agents on each task is more time-consuming. Excluding tasks that did not generate executable code, \Cref{fig:results-long-run} shows 200 iterations of \ours{} with the R2D2 robot. Each node represents a generated task, which includes the natural language task description and environment code. The color of the nodes corresponds to the generation number. For better visualization, each task (natural language description and environment code) is encoded using a pre-trained encoder language model (OpenAI's text-embedding-3-small~\citep{openai_text_embedding_3_small}) and then projected into a 2-dimensional space using t-SNE~\citep{van2008visualizing}. We manually selected a few tasks that are well-distributed across the embedding space to show a sample of the diversity and creativity of \ours{}. Connections between nodes indicate which tasks were used as stepping stones to generate the next task, meaning they were provided in context to the task generator. For better readability, \Cref{fig:results-long-run} only displays the parent task that is closest to the child task in the embedding space and the header text of task descriptions. \Cref{appendix:long-run} shows the full natural language descriptions of the magnified tasks and provides a more comprehensive visualization of all parent-child task connections. \Cref{appendix:task-desc-seeds} shows the three task descriptions that seed the run.

\ours{} creates tasks that evolve and diverge across the embedding space, forming clusters of related and increasingly complex challenges (\Cref{fig:results-long-run}). For example, the bottom-right region features tasks involving kicking a ball into a goal. Moving towards the bottom-left, the tasks gradually include dynamic elements such as moving obstacles or targets, while still focusing on kicking a ball. As we traverse to the top-left region, the focus shifts to tasks that require pushing or delivering objects into designated receptacles or target places. This marks a significant departure from the ball-kicking tasks and showcases the diversity of the generated environments. The top-right portion of the space is characterized by generations that emphasize navigation challenges, such as traversing (often moving) platforms or navigating across varied terrains.

Furthermore, there are niches where some generations appear to be variations of each other (\Cref{fig:results-long-run}). For example, in the top-right corner, the generations surrounding task 90 involve retrieving an object and returning it to a designated location. While the overarching objective surrounding that niche remains consistent, the tasks and their learning environments exhibit notable differences. These variations manifest in several aspects, such as the simulated world in which the task takes place (e.g., task 90 is outdoors, task 24 is in connected rooms, task 27 is in two levels connected by a ramp). Other variations include the number of objects that need to be retrieved (e.g., task 102 requires multiple objects to be retrieved, while task 90 only requires one), and the allotted time for completing the task (e.g., task 27 has a time limit of 3 minutes, while task 24 has a time limit of 5 minutes). These task variations can potentially allow for the development of robust and adaptable agents capable of handling a wide range of scenarios~\citep{bauer_HumanTimescaleAdaptationOpenEnded_2023}.

Overall, \ours{} generates tasks that significantly diverge from the seed tasks used to initialize the archive (the grey nodes in \Cref{fig:results-long-run}). For example, despite the absence of dynamic objects in the seed tasks, \ours{} generates a substantial number of tasks that incorporate dynamic objects and interactions, such as platforms that move horizontally and vertically, buttons and levers to activate, and moving obstacles. \ours{} not only explores different task niches (e.g., navigating across different terrains vs. retrieving objects) but also generates interesting variations within each niche (e.g., retrieving objects in different simulated world settings).

\vspace{-4pt}
\section{Short Run with Learning}
\label{sec:short-run}
\vspace{-4pt}

\begin{figure}[ht]
    \centering
    \includegraphics[width=\textwidth]{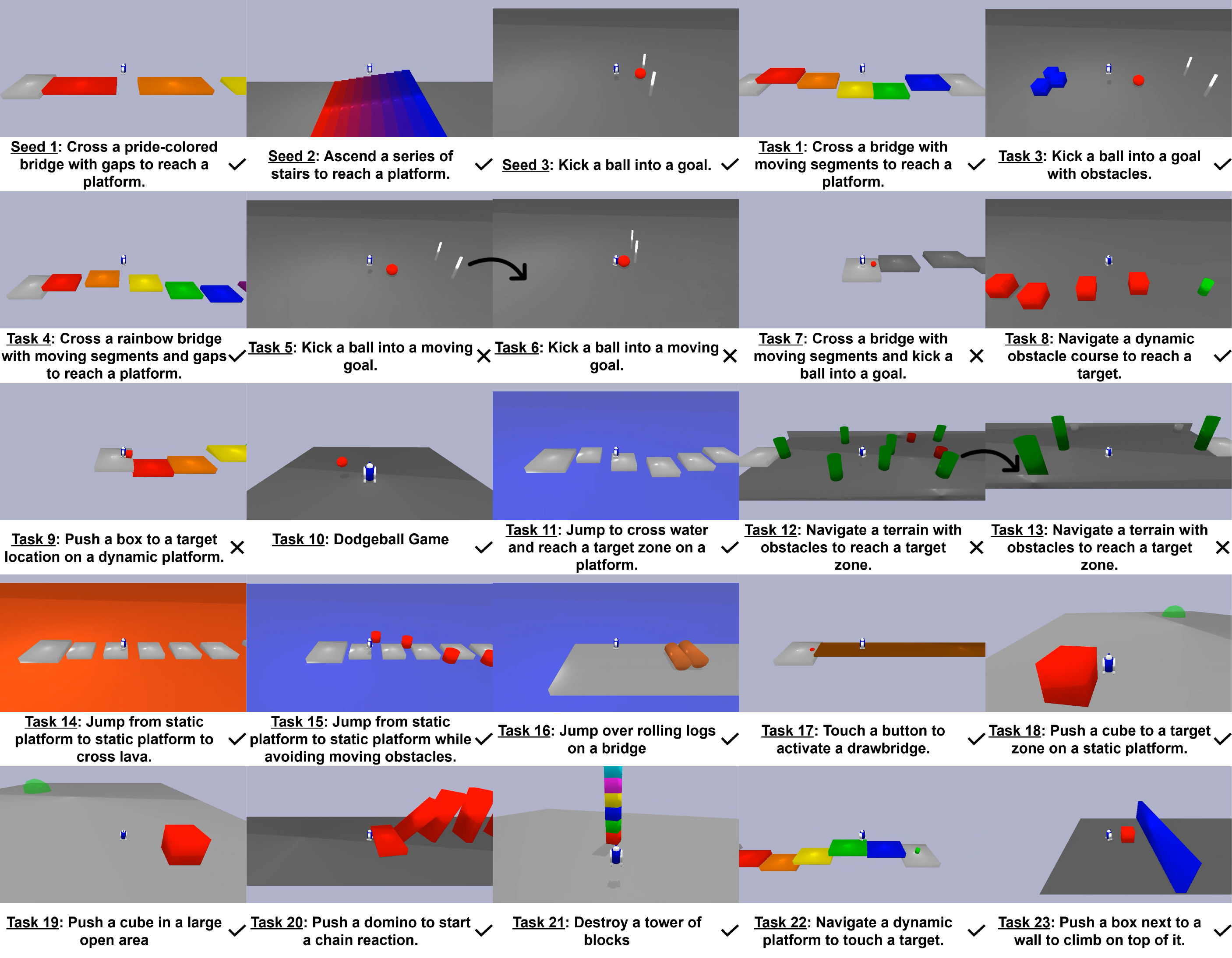}
    \vspace{-20pt}
    \caption{\textbf{Short Run with Learning.} \ours{} adapts to the current capabilities of trained RL agents, generating tasks that are both interesting and learnable. Tasks deemed interesting that are successfully learned are marked by a check and failures by a cross. Uninteresting tasks are not trained on and hence not included here. Arrows between tasks indicate instances where \ours{} modified a task that the RL agent failed to learn, adjusting the task difficulty to facilitate learning.}
    \label{fig:results-short-run}
    \vspace{-5pt}
\end{figure}

To demonstrate \ours{}'s ability to generate tasks of suitable difficulty for training RL agents, we conducted 5 short runs with RL agent training. Due to limited computational resources in our academic lab, the runs are shorter, but still show the creative potential of \ours{} and its ability to tailor tasks to the agents' abilities. The success detector (\Cref{subsec:success-detector}) evaluates if the agent has successfully completed each task. All short runs are initialized with 3 task description seeds (\Cref{appendix:task-desc-seeds}). We find that \ours{} can effectively generate tasks and environment code that are not only creative and interesting but also learnable and appropriately challenging for RL agents (\Cref{fig:results-short-run}, \Cref{appendix:more-shortruns}). In the run shown in \Cref{fig:results-short-run}, the RL agents successfully completed 16 tasks, failed at 6 tasks, and 1 task was deemed uninteresting by the post-generation MoI. We conducted a user study with 50 participants and found a 72.7\% alignment rate between human evaluations and the success detector's assessments (\Cref{appendix:humaneval-succdet}). \Cref{fig:results-short-run} shows images of the trained agents and their corresponding tasks, displaying only the header text of the task descriptions for better readability. \Cref{appendix:trained-agents} shows the environment code with full task descriptions, images, and a task graph.

\ours{} leverages previously learned tasks as stepping stones to generate and master more challenging tasks. This iterative process allows RL agents to build upon existing skills to tackle increasingly complex environments. For example, since RL agents in the archive successfully learned to cross a pride-colored bridge with gaps (seed 1) and cross a bridge with moving segments (task 1), \ours{} branched off these tasks to generate a more challenging task (task 4) combining these elements. This new task required the agent to cross a rainbow bridge with gaps and moving segments. By continuing training from the policy learned on task 1, the RL agent successfully completed task 4.
Furthermore, \ours{} considers tasks that the RL agents agents failed to learn. For example, since the RL agent failed to learn how to push a box to a target location on a dynamic platform (task 9), future tasks involving crossing platforms did not include pushing objects across them (e.g., tasks 11, 14, 15). This ensures that the generated tasks remain learnable and do not repeatedly incorporate overly difficult challenges.
Similarly, when the RL agent failed to navigate through a terrain with obstacles (task 12), \ours{} generates an easier obstacle course (task 13) that maintains the same objective but reduces the number of obstacles.
By using the agent's past experiences (both successes and failures) as building blocks, \ours{} generates a curriculum of tasks that progressively and interestingly increases in difficulty while remaining learnable (additional ablations in \Cref{appendix:shortrun-ablation}). This adaptive task generation process showcases \ours{}'s capacity to create a tailored curriculum that maintains an appropriate level of challenge, ensuring that the generated tasks are neither too simple nor too complex.

\vspace{-8pt}
\section{Quantitative Results}
\vspace{-8pt}

\begin{figure}[ht]
    \centering
    \includegraphics[width=0.45\textwidth]{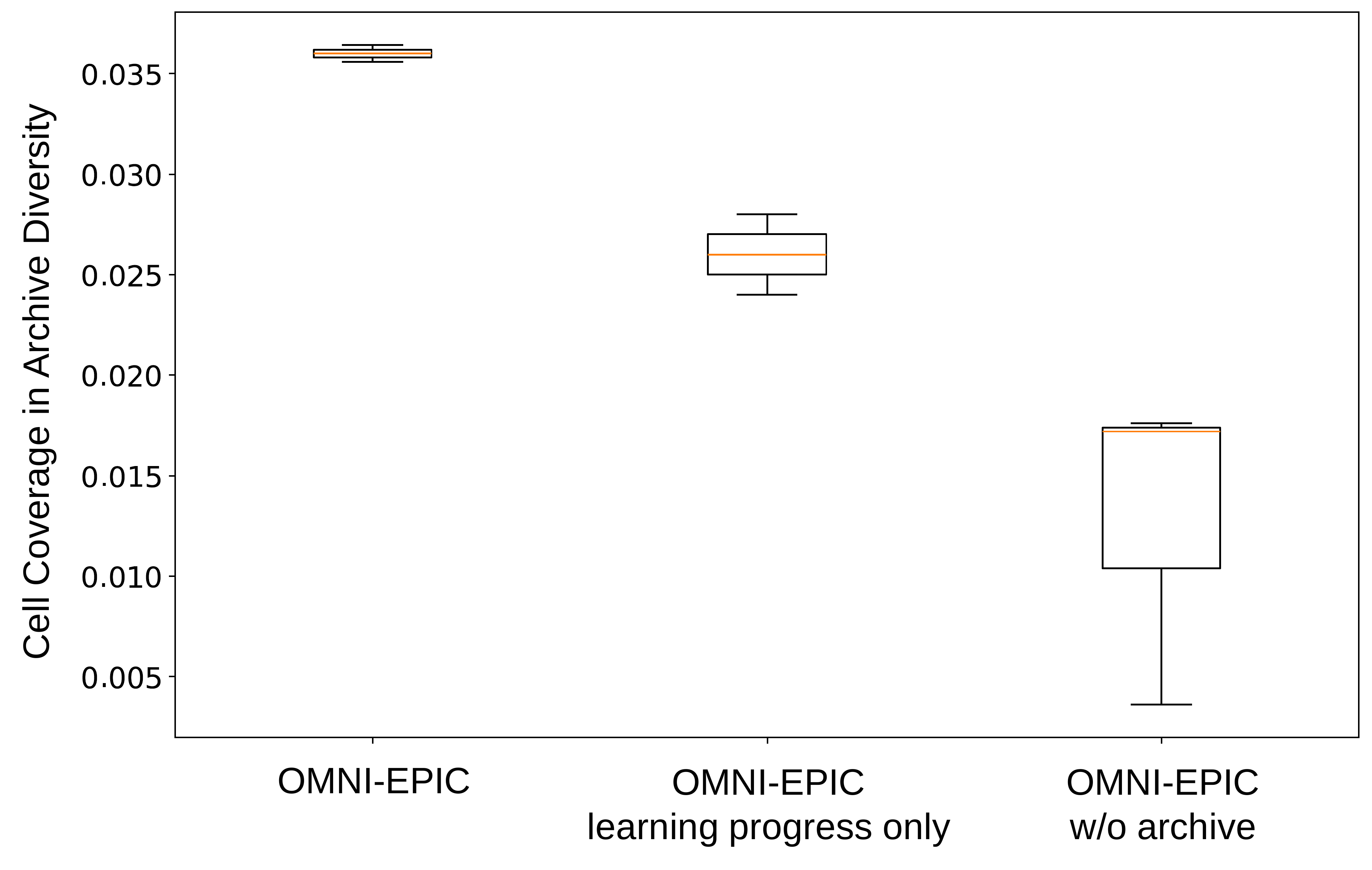}
    \includegraphics[width=0.45\textwidth]{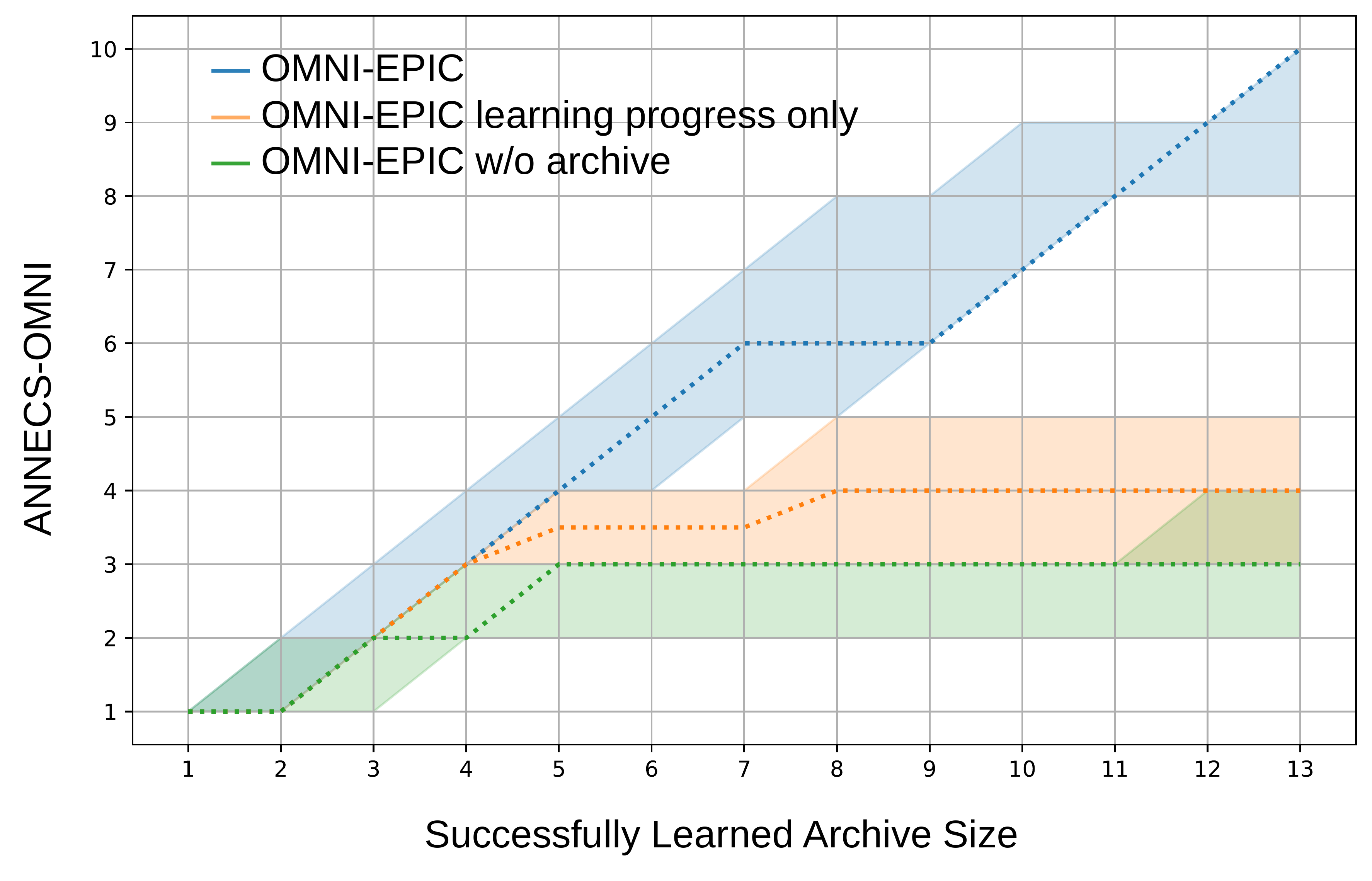}
    \vspace{-10pt}
    \caption{\textbf{OMNI-EPIC generates significantly more diverse tasks and continues to innovate throughout the run.} (Left) Cell coverage of archive diversity plots in long runs with simulated learning by OMNI-EPIC and the controls. (Right) ANNECS-OMNI measure of progress for OMNI-EPIC and the controls. Dotted lines are median values, shaded regions are 95\% confidence intervals.}
    \label{fig:quantitative}
    \vspace{-2pt}
\end{figure}

To evaluate the impact of having a task archive and the contribution of the notion of interestingness in OMNI-EPIC, we compare against two controls: (1) OMNI-EPIC without the task archive (\textbf{OMNI-EPIC w/o archive}), and (2) OMNI-EPIC without the models of interestingness (by removing the request for interesting tasks from the task generator prompts and skipping the post-generation MoI step) (\textbf{OMNI-EPIC Learning Progress Only}). The quantitative measures for comparisons are described next, and OMNI-EPIC significantly outperforms the controls on both metrics.

\textbf{Task Diversity.}
To measure the diversity of generated task archives, we plot the cell coverage of a task archive in a 2D discretized plot. First, we use a pretrained text embedding model (OpenAI's text-embedding-3-small~\citep{openai_text_embedding_3_small}) to encode the generated tasks (natural language description and environment code), then reduce the dimensionality to two via PCA~\citep{mackiewicz1993principal} (across all tasks from all methods) for easier visualization. We create 2D discretized plots by selecting a discretization level (e.g., 50 in \Cref{fig:quantitative}) and generating uniform bins across the minimum and maximum values from the PCA embeddings. Each task in the archive is placed in the appropriate bin, and we count the number of unique bins the algorithm fills, a standard measure of diversity~\citep{mouret2015illuminating, pugh2016quality}. Each method is run with simulated learning, repeated 3 times. OMNI-EPIC achieves significantly higher cell coverage than the controls (p < 0.05, Mann-Whitney U test). This shows that both the task archive and the MoI significantly and quantifiably contribute to OMNI-EPIC's ability to generate more diverse tasks (\Cref{fig:quantitative}). \Cref{appendix:quantitative} shows the archive diversity plots for each method and that the same results hold for different discretization levels.

\textbf{Measure of Progress.}
While open-ended systems aim to endlessly create and learn new tasks, the question of how to measure progress in such systems remains. \citet{wang_EnhancedPOETOpenended_2020} proposed tracking the Accumulated Number of Novel Environments Created and Solved (ANNECS) throughout the run of an open-ended system. Specifically, ANNECS requires that an environment created at a particular iteration (1) is not too hard or easy for the agent to learn and (2) must eventually be solved by the system. However, ANNECS lacks a measure of how interesting or novel the newly generated task is. To address this limitation, we introduce a new metric to measure progress in open-ended systems, ANNECS-OMNI. Inspired by ideas in \citet{zhang_OMNIOpenendednessModels_2023}, ANNECS-OMNI adds a third criterion to ANNECS: the new task must be considered interesting compared to previous tasks (approximated here by asking an FM if the task is interesting given the archive of already-solved tasks).

Each method is run with RL training, repeated 5 times. OMNI-EPIC achieves significantly higher ANNECS-OMNI scores than the controls (p < 0.05, Mann-Whitney U test). As the run proceeds, the ANNECS-OMNI metric consistently increases for OMNI-EPIC, indicating that, for as long as we could afford to run it, the algorithm continuously creates meaningfully new and interesting tasks without stagnation (\Cref{fig:quantitative}). That is a new high watermark in our field’s longstanding quest to create open-ended algorithms. The ANNECS metric for the different methods can be found in \Cref{appendix:quantitative}.

\vspace{-7pt}
\section{Discussion, Future Work, Conclusion}
\label{sec:conclusion}
\vspace{-7pt}
\ours{} provides a general recipe for generating a potentially endless stream of learnable and interesting environments. By leveraging FMs to generate tasks and their corresponding environment (and reward) code, \ours{} eliminates the need for handcrafted parameters or predefined task distributions, unshackling algorithms to have the potential to be truly open-ended. \ours{} can produce a wide variety of tasks, from simple to complex, and allows for the creation of environments that continuously adapt in response to an agent's developing capabilities. The results demonstrate its effectiveness in generating diverse and creative tasks, highlighting the potential of this approach for training more capable and intelligent agents. By leveraging the generative capabilities of FMs to create a vast array of unique and challenging environments, \ours{} brings us one step closer to achieving Darwin Completeness, the ability to create \emph{any} possible learning environment.

While we cannot rule out that \ours{} is creating environments similar to those in its training data, (1) even if it were, this would still be useful for open-endedness; (2) we believe it is generating novel environments, as we were unable to find the same environments when searching online; and (3) with longer runs, we hypothesize it could generate environments endlessly. Its task generator and MoI could generalize well outside the distribution of human data when conditioned on a growing archive of discoveries, though studying this remains a fascinating area for future research.

However, it is important to acknowledge that the current implementation of \ours{} is not yet Darwin Complete, primarily due to the limitations imposed by the choice of simulator. While \ours{} can generate a wide variety of tasks and environments, it is constrained by the capabilities and assumptions of the underlying simulation platform. To achieve true Darwin Completeness, \ours{} could simply be allowed to generate any code, including the ability to download, install, use, or modify any existing simulator, or even write the code for an entirely new simulator from scratch.
Since the programming language used here (Python) is Turing Complete, generating code can, in principle, create any computable environment.
Enabling the generation of arbitrary code would unlock the full potential of \ours{} and bring us closer to realizing the vision of Darwin Completeness. Of course, ever smarter FMs are required to take advantage of this opportunity.

The current implementation of \ours{} trains a population of specialist agents. We do not provide evidence of their generalization across tasks as it is not the focus of the paper. Future research could explore alternative training modalities. One promising direction is to train a single policy across all environments~\citep{bauer_HumanTimescaleAdaptationOpenEnded_2023}, which could encourage the development of more versatile and adaptable agents (\Cref{appendix:generalists}). Another approach is to prioritize environments based on learning progress~\citep{baranes2013active, kanitscheider2021multi}, focusing on tasks that provide the most opportunities for improvement. Each of these strategies introduces unique dynamics into the open-ended environment generation process, and understanding their effects on agent performance and generalization represents an exciting avenue for future research.

In conclusion, \ours{} represents a leap towards open-ended learning by generating a potentially endless stream of learnable and interesting tasks. Intriguingly, it also provides a new way of creating human entertainment and educational resources by offering a limitless supply of engaging challenges (\Cref{appendix:game-interface}). \ours{} could potentially be applied in myriad ways, covering anything from math problems and poetry challenges to games and virtual worlds. By leveraging FMs to create tasks and environment code, \ours{} opens up a vast space of possibilities for AI and human agents to explore and master. By combining that expressive power with human notions of interestingness, \ours{} presents a promising path towards the development of truly open-ended and creative AI.

\subsubsection*{Acknowledgments}
This research was supported by the Vector Institute, the Canada CIFAR AI Chairs program, a grant from Schmidt Futures, an NSERC Discovery Grant, the Center for AI Safety Compute Cluster, DARPA and a generous donation from Rafael Cosman. Any opinions, findings, and conclusions or recommendations expressed in this material are those of the authors and do not necessarily reflect the views of the sponsors. We also thank Aaron Dharna, Arthur Braida, Ben Norman, Cong Lu, Gabriel Béna, Luca Grillotti, Rach Pradhan, and Shengran Hu, for insightful discussions and feedback.

\bibliography{main}
\bibliographystyle{iclr2025_conference}

\appendix
\newpage

\section{Reproducibility}
For full transparency and to ensure reproducibility, we have open-sourced all code associated with this work. The code can be found at \url{https://github.com/maxencefaldor/omni-epic}.

\section{Game Interface}
\label{appendix:game-interface}

\begin{figure}[ht]
    \centering
    \includegraphics[width=\linewidth]{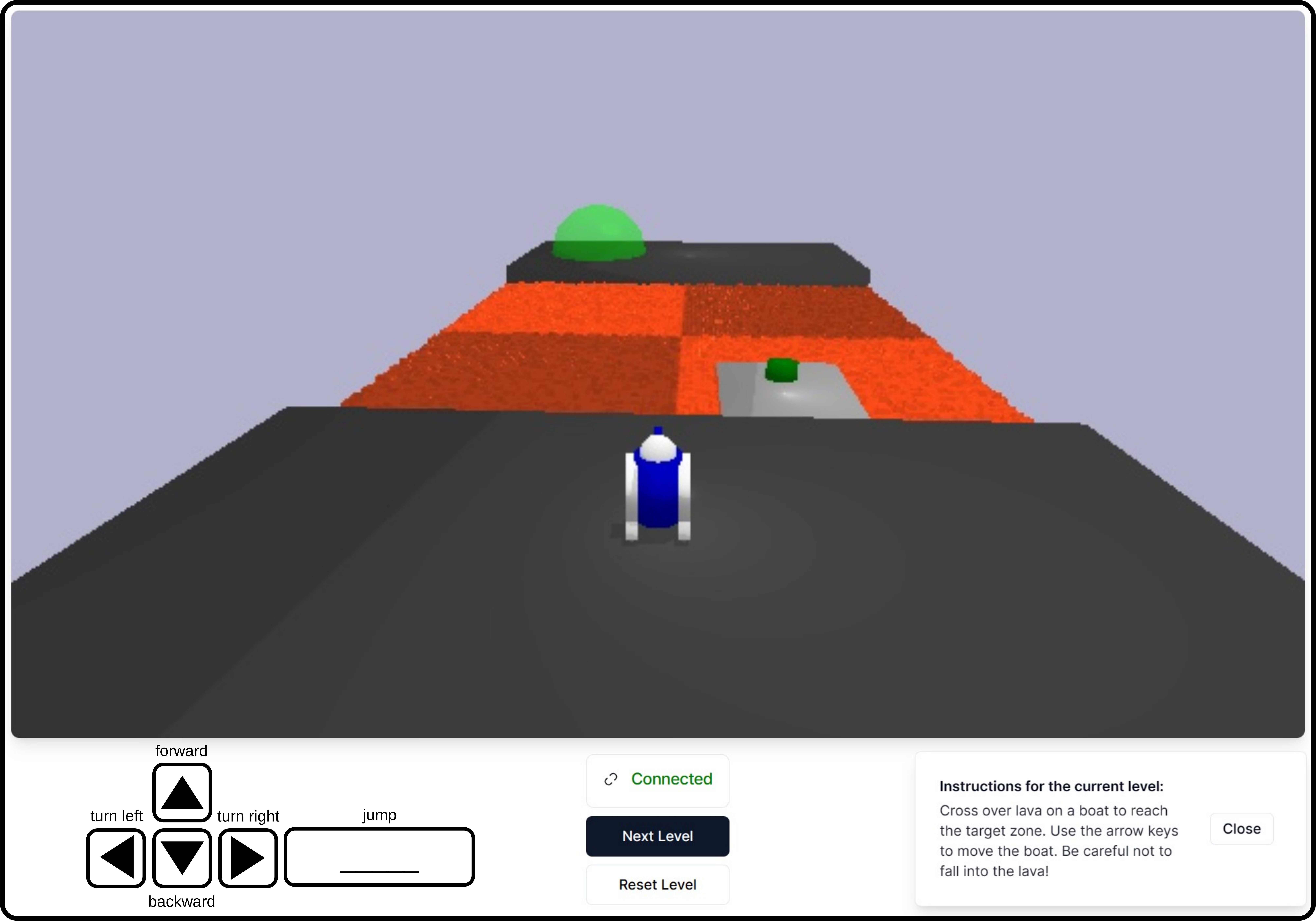}
    \caption{\textbf{\ours{} in a game interface.}}
    \label{fig:game}
\end{figure}

Apart from training agents, \ours{} can also be a good form of human entertainment! We created a game interface whereby players can control the R2D2 robot with keyboard inputs, and play the generated tasks (\Cref{fig:game}). \ours{} dynamically generates the next interesting level for the player, adjusting based on the player's skill level by suggesting tasks that are not too easy or difficult. \ours{} opens a new era for games, where procedural content is automatically generated and tailored to the player's abilities, ensuring a consistently engaging experience.

\newpage
\section{VLM as Success Detector}
\label{appendix:vlm-succdet}

The success-checking function can be easily implemented in simulated environments where information can be readily accessed through code (\Cref{subsec:success-detector}). However, it may face challenges in real-world scenarios, off-the-shelf closed-source video games, or even within simulated environments for tasks that involve visual assessment (e.g., building a castle, arranging boxes to resemble an elephant), where the required information is not directly available or difficult to evaluate using only code. A natural solution is to use VLMs as success detectors~\citep{radford2021learning}. VLMs can potentially detect success on a wider range of tasks (e.g., tidying a room, doing a backflip)~\citep{du2023vision} than code generation. We input snapshots of the agent's behavior every second, the natural language task description, and environment code (see below).
Since our preliminary testing found that current VLMs are not yet accurate enough to be used as success detectors, we use code generated by LLMs for this purpose instead. However, we expect VLM capabilities to rapidly improve over time, eventually achieving higher accuracy and making them viable for future use.

\textbf{System Prompt:}
\begin{lstlisting}[language={}]
You are an expert in Python programming and reinforcement learning. Your goal is to evaluate if a robot has solved a task. You will be provided with the task description, the corresponding environment code and an image containing snapshots of the robot attempting to complete the task. Your objective is to describe the image, reason about whether the task has been completed and determine if the robot has solved the task.

Instructions:
- In the description of the image, describe the environment and the behavior of the robot.
- In the reasoning, analyze if the environment corresponds to the task description and if the behavior of the robot meets the requirements for task success.
- The task is considered failed if the environment is constructed in a way that makes solving the task impossible.
- If you are unsure, make an educated guess and always provide an answer.
- If you are unsure, say that it has failed.

Robot description:
{ROBOT_DESC}

Desired format:
Description of the image:
<image description>

Reasoning for task success/failure:
<reasoning>

Did the robot solve the task?:
<Yes/No>
\end{lstlisting}

\textbf{User Prompt:}
\begin{lstlisting}[language={}]
Task description and environment code:
{ENV_CODE}
\end{lstlisting}

\textbf{Image Examples:}
\begin{figure}[ht]
    \centering
    \includegraphics[width=\textwidth]{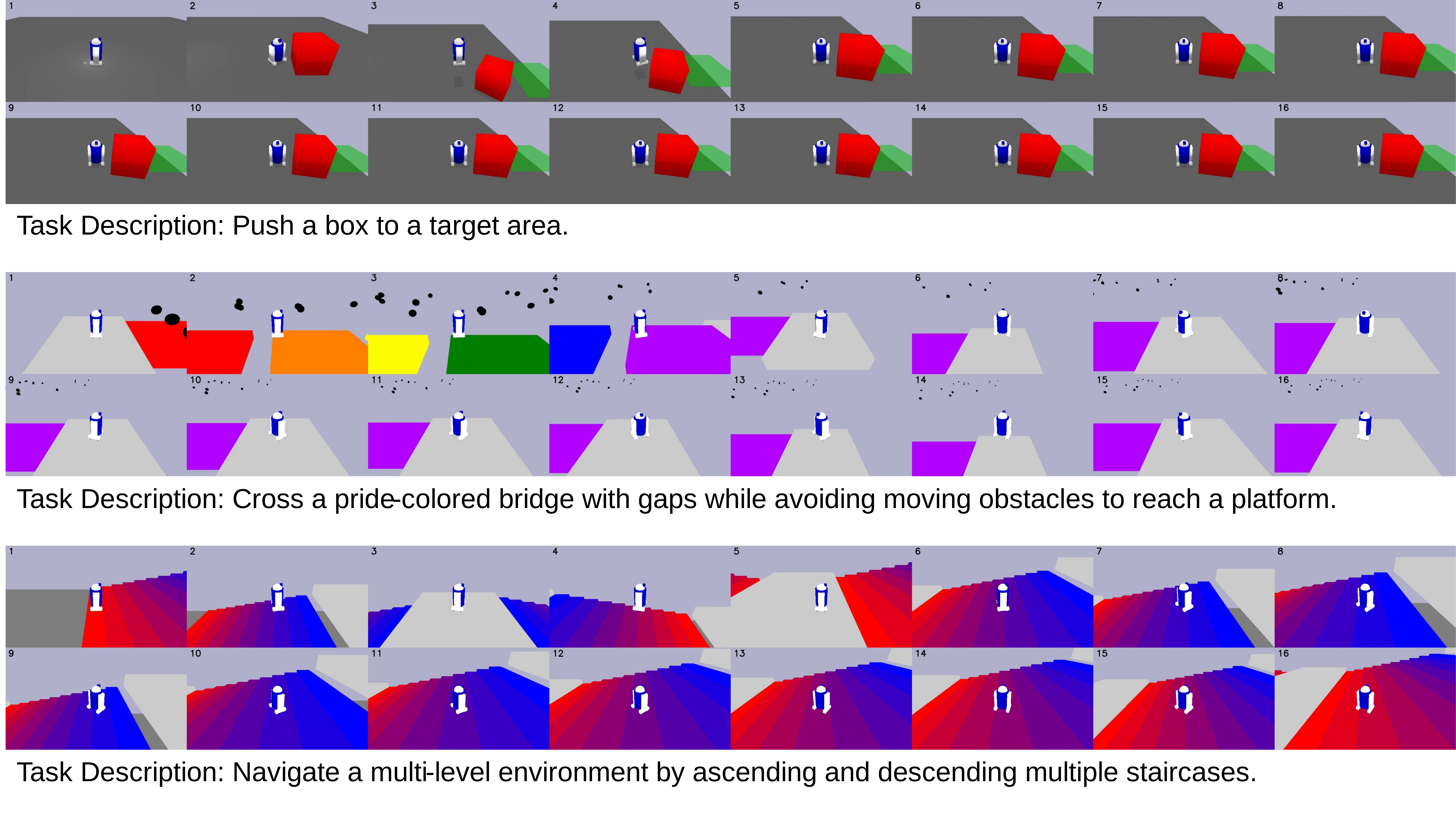}
\end{figure}

\newpage
\section{Supplementary Materials for Long Run with Simulated Learning}
\label{appendix:long-run}

\begin{figure}[ht]
    \centering
    \includegraphics[width=\textwidth]{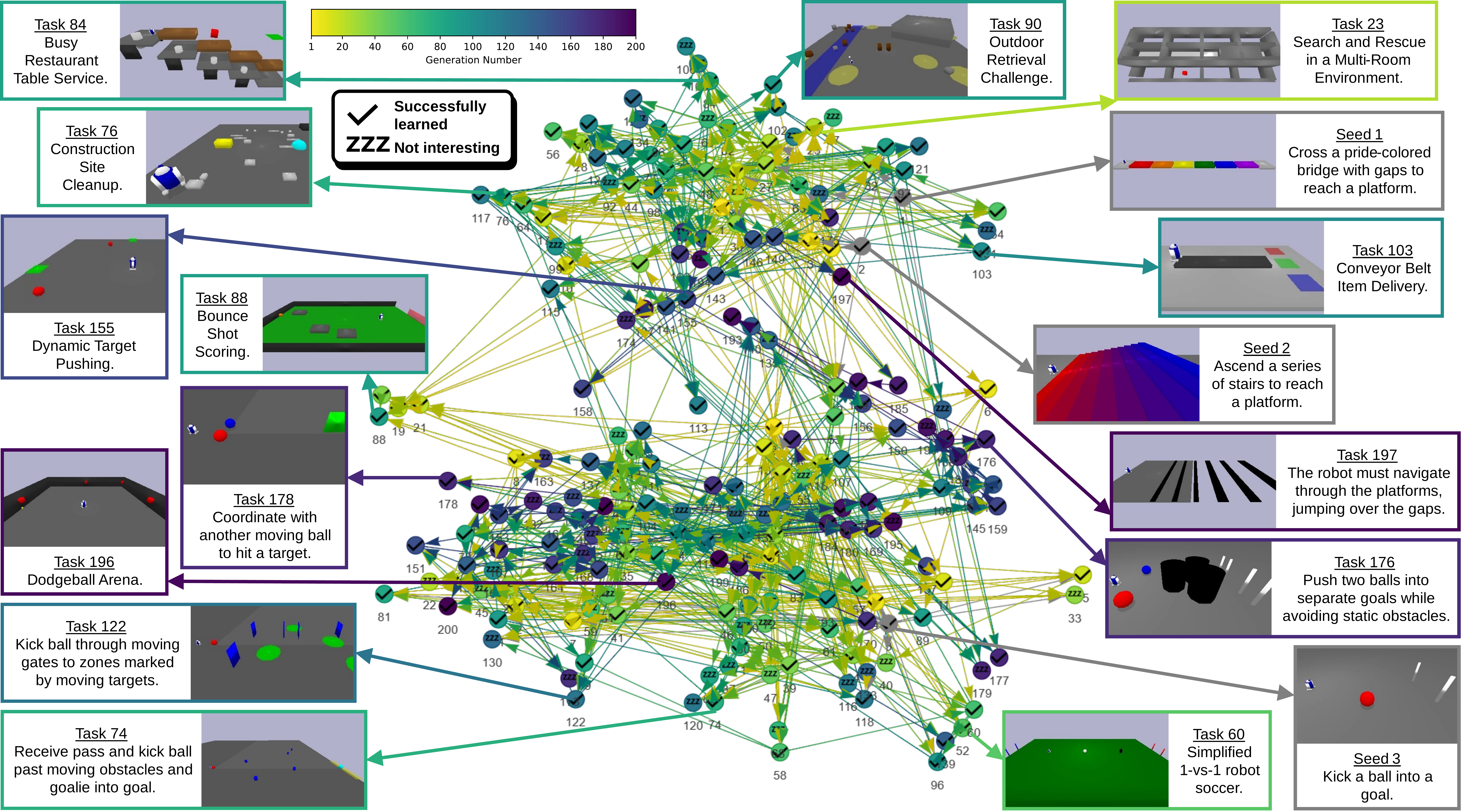}
    \caption{\textbf{Long run with simulated learning task graph with all parent-child task connections.} This figure presents the same task graph as \Cref{fig:results-long-run}, but with all parent-child task connections displayed. The node color reflects the generation number of the task. A check mark in the node means that the task was successfully learned. A ZZZ symbol means that the task was deemed uninteresting and discarded. The node connections illustrate which tasks were conditioned on when asking an FM to generate a similar yet new and interesting task. Due to the high density of tasks and connections, visualizing all relationships clearly in a static image is challenging. To better understand and navigate the intricate web of task relationships, an interactive version of the task graph is available at \website{}.}
    \label{fig:results-long-run-all}
\end{figure}

% \begin{figure}[ht]
%     \centering
%     \includegraphics[width=\textwidth]{figures/results_longrun_graynodes.pdf}
%     \caption{\textbf{Long run with simulated learning task graph where only the nodes with enlarged task images are colored.} This figure presents the same task graph as \Cref{fig:results-long-run}, but all nodes are grayed out except for those with enlarged task images, which are colored.}
% \end{figure}

% \begin{figure}[ht]
%     \centering
%     \includegraphics[width=\textwidth]{figures/results_longrun_thinneredges.pdf}
%     \caption{\textbf{Long run with simulated learning task graph with thinner visualization edges.} This figure presents the same task graph as \Cref{fig:results-long-run}, but the edges are visualized with reduced thickness.}
% \end{figure}

The natural language descriptions and environment code for all tasks shown in \Cref{fig:results-long-run} are available at \website{}. For conciseness, we present the full natural language descriptions of all magnified tasks, and the environment code of one magnified task (Task 76):
\begin{lstlisting}[language=Python]
import numpy as np
from oped.envs.r2d2.base import R2D2Env

class Env(R2D2Env):
    """
    Construction Site Cleanup

    Description: 
    - The environment is a construction site with dimensions 10 m x 10 m.
    - There are 3 piles of different types of debris scattered around the construction site:
      - 5 bricks (each 0.2 m x 0.2 m x 0.2 m)
      - 5 metal scraps (each 0.5 m x 0.25 m x 0.25 m) 
      - 5 wooden planks (each 1 m x 0.2 m x 0.1 m)
    - There are also 2 robotic construction vehicles moving around the site in pre-programmed patterns:
      - A bulldozer that pushes dirt piles around in a rectangle, 4 m x 3 m
      - An excavator that swings its arm and bucket back and forth in a 180 degree arc, 2 m in radius
    - On one side of the construction site are 3 square receptacle bins, each 1 m x 1 m x 0.5 m tall, labeled for each type of debris (bricks, metal, wood).
    - The robot starts in one corner of the site.

    Task:
    The robot needs to pick up each piece of debris and place it in the correct receptacle, while avoiding collisions with the moving construction vehicles. The robot can only carry one piece of debris at a time.

    Success Conditions:
    The task is complete when all pieces of debris have been placed in their correct bins, and the robot has returned to its starting position.

    Time Limit:
    The robot has 10 minutes to complete the cleanup task.

    Rewards:
    - Provide a small reward for each piece of debris successfully picked up.
    - Provide a moderate reward for each piece of debris placed in the correct bin.
    - Provide a large reward for completing the task and returning to the start position.
    - Provide a small penalty for each collision with the construction vehicles.

    Termination:
    The episode ends if the robot flips over, or if the time limit is exceeded.
    """

    def __init__(self):
        super().__init__()
        self.site_size = [10.0, 10.0, 0.1]
        self.site_position = [0.0, 0.0, 0.0]
        self.site_id = self.create_box(mass=0.0, half_extents=[self.site_size[0] / 2, self.site_size[1] / 2, self.site_size[2] / 2], position=self.site_position, color=[0.5, 0.5, 0.5, 1.0])
        self._p.changeDynamics(bodyUniqueId=self.site_id, linkIndex=-1, lateralFriction=0.8, restitution=0.5)
        self.brick_size = [0.2, 0.2, 0.2]
        self.metal_size = [0.5, 0.25, 0.25]
        self.wood_size = [1.0, 0.2, 0.1]
        self.debris_sizes = [self.brick_size, self.metal_size, self.wood_size]
        self.num_debris_types = len(self.debris_sizes)
        self.num_debris_each = 5
        self.debris_ids = []
        for i in range(self.num_debris_types):
            for _ in range(self.num_debris_each):
                debris_position = [np.random.uniform(-self.site_size[0] / 2 + 2.0, self.site_size[0] / 2 - 2.0), np.random.uniform(-self.site_size[1] / 2 + 2.0, self.site_size[1] / 2 - 2.0), self.debris_sizes[i][2] / 2]
                debris_id = self.create_box(mass=1.0, half_extents=[self.debris_sizes[i][0] / 2, self.debris_sizes[i][1] / 2, self.debris_sizes[i][2] / 2], position=debris_position, color=[0.8, 0.8, 0.8, 1.0])
                self.debris_ids.append((debris_id, i))
        self.bulldozer_size = [1.0, 0.5, 0.5]
        self.bulldozer_position_init = [-self.site_size[0] / 4, 0.0, self.bulldozer_size[2] / 2]
        self.bulldozer_id = self.create_box(mass=0.0, half_extents=[self.bulldozer_size[0] / 2, self.bulldozer_size[1] / 2, self.bulldozer_size[2] / 2], position=self.bulldozer_position_init, color=[1.0, 1.0, 0.0, 1.0])
        self.excavator_radius = 2.0
        self.excavator_position_init = [self.site_size[0] / 4, 0.0, 0.0]
        self.excavator_id = self.create_sphere(mass=0.0, radius=0.5, position=self.excavator_position_init, color=[0.0, 1.0, 1.0, 1.0])
        self.receptacle_size = [1.0, 1.0, 0.5]
        self.receptacle_positions = [[self.site_size[0] / 2 - self.receptacle_size[0] / 2, -self.site_size[1] / 3, self.receptacle_size[2] / 2], [self.site_size[0] / 2 - self.receptacle_size[0] / 2, 0.0, self.receptacle_size[2] / 2], [self.site_size[0] / 2 - self.receptacle_size[0] / 2, self.site_size[1] / 3, self.receptacle_size[2] / 2]]
        self.receptacle_ids = []
        for i in range(self.num_debris_types):
            receptacle_id = self.create_box(mass=0.0, half_extents=[self.receptacle_size[0] / 2, self.receptacle_size[1] / 2, self.receptacle_size[2] / 2], position=self.receptacle_positions[i], color=[0.2, 0.2, 0.2, 1.0])
            self.receptacle_ids.append(receptacle_id)
        self.robot_position_init = [-self.site_size[0] / 2 + 2.0, -self.site_size[1] / 2 + 2.0, self.site_size[2] + self.robot.links['base'].position_init[2] + 0.1]
        self.robot_orientation_init = self._p.getQuaternionFromEuler([0.0, 0.0, np.pi / 4])
        self.time_limit = 600.0
        self.debris_pick_reward = 1.0
        self.debris_place_reward = 10.0
        self.task_complete_reward = 100.0
        self.collision_penalty = -1.0

    def create_box(self, mass, half_extents, position, color):
        collision_shape_id = self._p.createCollisionShape(shapeType=self._p.GEOM_BOX, halfExtents=half_extents)
        visual_shape_id = self._p.createVisualShape(shapeType=self._p.GEOM_BOX, halfExtents=half_extents, rgbaColor=color)
        return self._p.createMultiBody(baseMass=mass, baseCollisionShapeIndex=collision_shape_id, baseVisualShapeIndex=visual_shape_id, basePosition=position)

    def create_sphere(self, mass, radius, position, color):
        collision_shape_id = self._p.createCollisionShape(shapeType=self._p.GEOM_SPHERE, radius=radius)
        visual_shape_id = self._p.createVisualShape(shapeType=self._p.GEOM_SPHERE, radius=radius, rgbaColor=color)
        return self._p.createMultiBody(baseMass=mass, baseCollisionShapeIndex=collision_shape_id, baseVisualShapeIndex=visual_shape_id, basePosition=position)

    def get_object_position(self, object_id):
        return np.asarray(self._p.getBasePositionAndOrientation(object_id)[0])

    def reset(self):
        observation = super().reset()
        self.time = 0.0
        self.debris_picked = [False] * len(self.debris_ids)
        self.debris_placed = [False] * len(self.debris_ids)
        self._p.resetBasePositionAndOrientation(self.robot.robot_id, self.robot_position_init, self.robot_orientation_init)
        return observation

    def step(self, action):
        observation, reward, terminated, truncated, info = super().step(action)
        self.time += self.dt
        bulldozer_position = self.get_object_position(self.bulldozer_id)
        bulldozer_position[0] = self.bulldozer_position_init[0] + 2.0 * np.sin(2 * np.pi * self.time / 20.0)
        bulldozer_position[1] = self.bulldozer_position_init[1] + 1.5 * np.sin(2 * np.pi * self.time / 30.0)
        self._p.resetBasePositionAndOrientation(self.bulldozer_id, bulldozer_position, [0.0, 0.0, 0.0, 1.0])
        excavator_position = self.get_object_position(self.excavator_id)
        excavator_position[0] = self.excavator_position_init[0] + self.excavator_radius * np.cos(np.pi * self.time / 10.0)
        excavator_position[1] = self.excavator_position_init[1] + self.excavator_radius * np.sin(np.pi * self.time / 10.0)
        self._p.resetBasePositionAndOrientation(self.excavator_id, excavator_position, [0.0, 0.0, 0.0, 1.0])
        return (observation, reward, terminated, truncated, info)

    def get_task_rewards(self, action):
        reward_pick = 0.0
        reward_place = 0.0
        reward_complete = 0.0
        penalty_collision = 0.0
        for i, (debris_id, debris_type) in enumerate(self.debris_ids):
            if not self.debris_picked[i] and len(self._p.getContactPoints(bodyA=self.robot.robot_id, bodyB=debris_id)) > 0:
                self.debris_picked[i] = True
                reward_pick += self.debris_pick_reward
            if self.debris_picked[i] and (not self.debris_placed[i]) and (len(self._p.getContactPoints(bodyA=debris_id, bodyB=self.receptacle_ids[debris_type])) > 0):
                self.debris_placed[i] = True
                reward_place += self.debris_place_reward
        if all(self.debris_placed) and np.linalg.norm(self.robot.links['base'].position[:2] - np.asarray(self.robot_position_init[:2])) < 1.0:
            reward_complete = self.task_complete_reward
        if len(self._p.getContactPoints(bodyA=self.robot.robot_id, bodyB=self.bulldozer_id)) > 0 or len(self._p.getContactPoints(bodyA=self.robot.robot_id, bodyB=self.excavator_id)) > 0:
            penalty_collision = self.collision_penalty
        return {'reward_pick': reward_pick, 'reward_place': reward_place, 'reward_complete': reward_complete, 'penalty_collision': penalty_collision}

    def get_terminated(self, action):
        if self.time >= self.time_limit:
            return True
        if np.dot(np.asarray([0, 0, 1]), np.asarray(self._p.getMatrixFromQuaternion(self.robot.links['base'].orientation)).reshape(3, 3)[:, 2]) < 0.5:
            return True
        return False

    def get_success(self):
        return all(self.debris_placed) and np.linalg.norm(self.robot.links['base'].position[:2] - np.asarray(self.robot_position_init[:2])) < 1.0
\end{lstlisting}

\foreach \i in {23, 60, 74, 84, 88, 90, 103, 122, 155, 176, 178, 196, 197} {
    \textbf{Task \i}\
    \input{task_descs/long_run/long_run_\i}
}

\newpage
\section{Supplementary Materials for Short Run with Learning}
\label{appendix:trained-agents}

\begin{figure}[ht]
    \centering
    \includegraphics[width=0.7\textwidth]{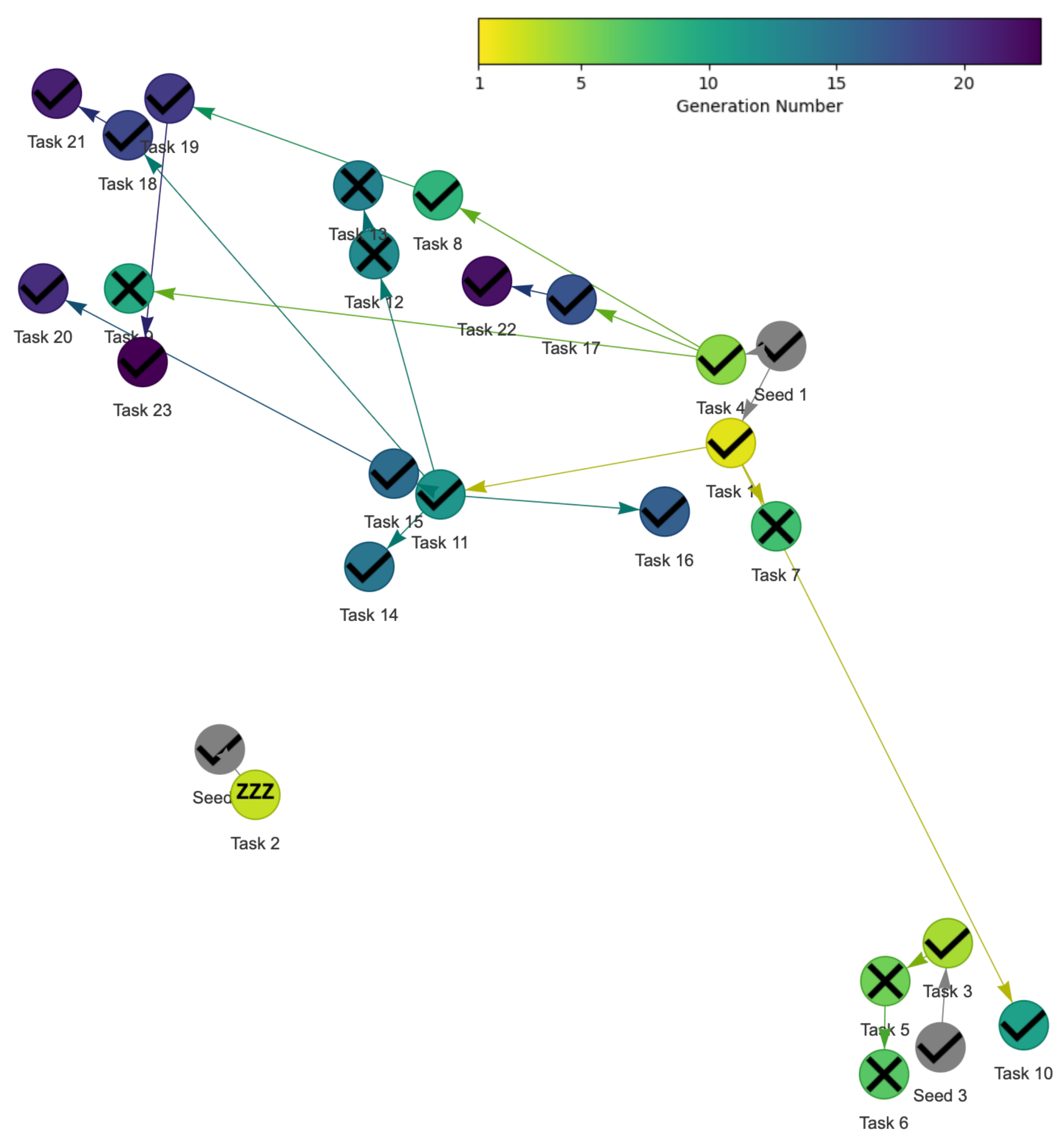}
    \caption{\textbf{Short run with Learning task graph.} The node color reflects the generation number of the task. A node with a check mark indicates successful task learning, while a cross mark denotes a task that was attempted but failed to be learned. A ZZZ symbol means that the task was deemed uninteresting and discarded. The node connections illustrate which tasks were conditioned on when asking an FM to generate a similar yet new and interesting task.}
\end{figure}

\foreach \i in {1, 3, 4, 5, 6, 7, 8, 9, 10, 11, 12, 13, 14, 15, 16, 17, 18, 19, 21, 23, 25} {
    \textbf{Task \i}\
    \begin{figure}[!ht]
        \centering
        \includegraphics[width=\textwidth]{figures/short_run/short_run_\i.png}
        \caption{Task \i\,of the \ours{} run presented in \Cref{sec:short-run}.}
    \end{figure}
    \input{env_codes/short_run/short_run_\i}
}

\newpage
\section{Additional Short Runs with Learning}
\label{appendix:more-shortruns}

Although the initial seed tasks in the task archive remain the same across repeated runs, we observe significant differences in the generated environments. This suggests that \ours{}'s ability to generate diverse learnable tasks is not heavily influenced by the initial set of environments.

\begin{figure}[h!]
    \centering
    \includegraphics[width=\textwidth]{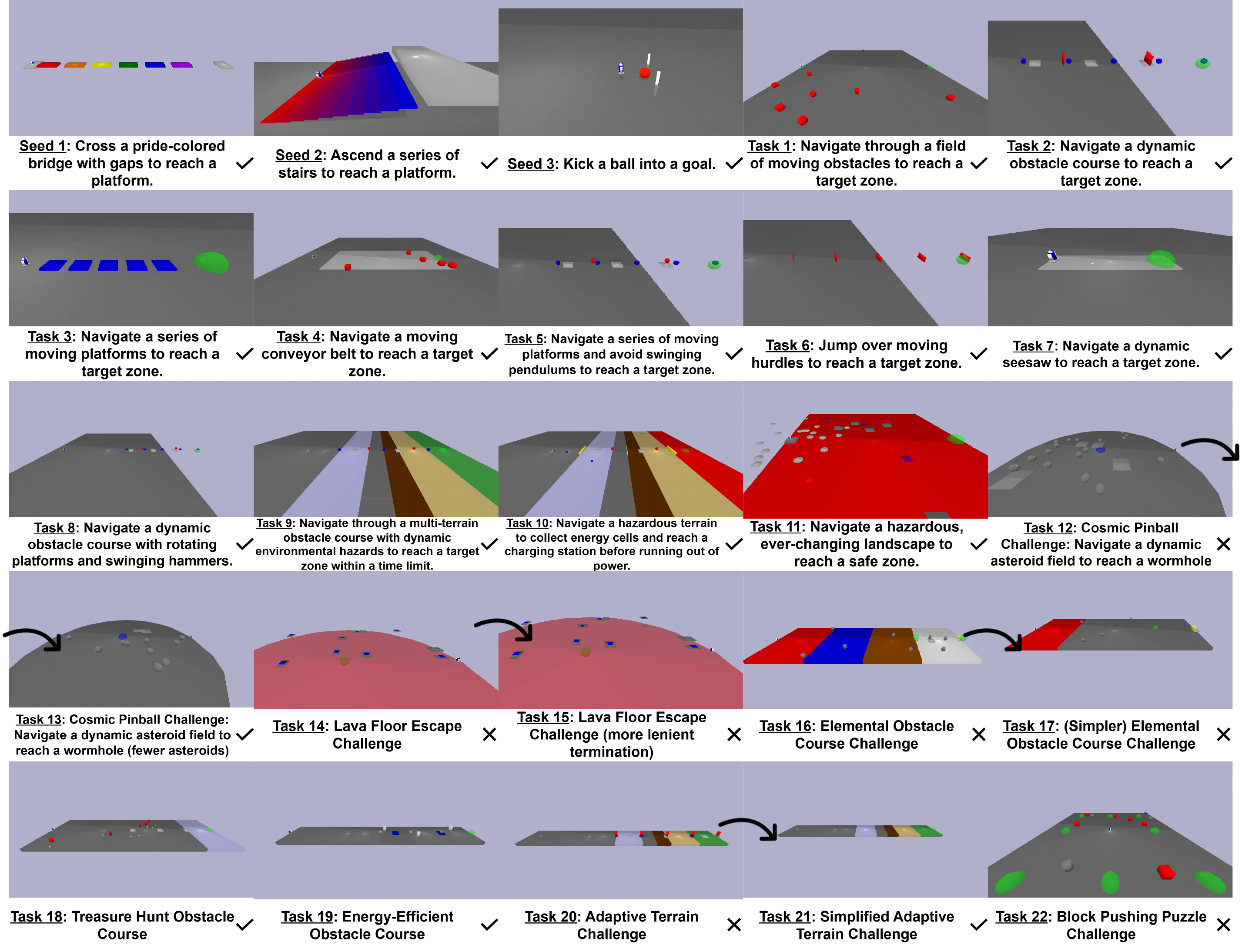}
    \caption{\textbf{Additional Short Run with Learning (Run 2).} This figure shows another instance of the short run experiment under the same settings as in \Cref{fig:results-short-run}. It demonstrates the progression of tasks generated by \ours{} and the learning outcomes of the RL agent. Checkmarks indicate successful learning, crosses indicate failures, and arrows show iterations on failed tasks.}
\end{figure}
\clearpage

\begin{figure}[h!]
    \centering
    \includegraphics[width=\textwidth]{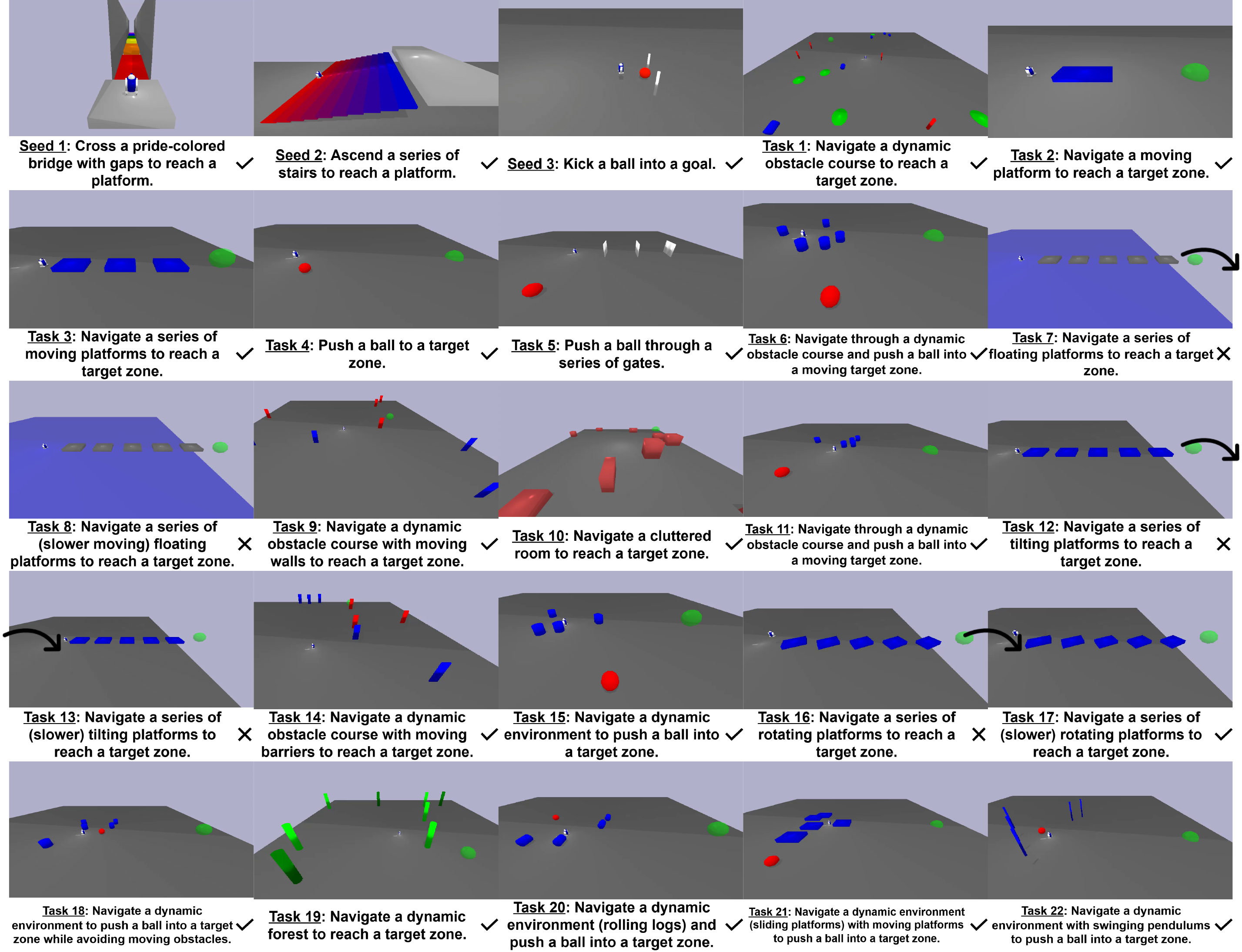}
    \caption{\textbf{Additional Short Run with Learning (Run 3).} This figure shows another instance of the short run experiment under the same settings as in \Cref{fig:results-short-run}. It demonstrates the progression of tasks generated by \ours{} and the learning outcomes of the RL agent. Checkmarks indicate successful learning, crosses indicate failures, and arrows show iterations on failed tasks.}
\end{figure}
\clearpage

\begin{figure}[h!]
    \centering
    \includegraphics[width=\textwidth]{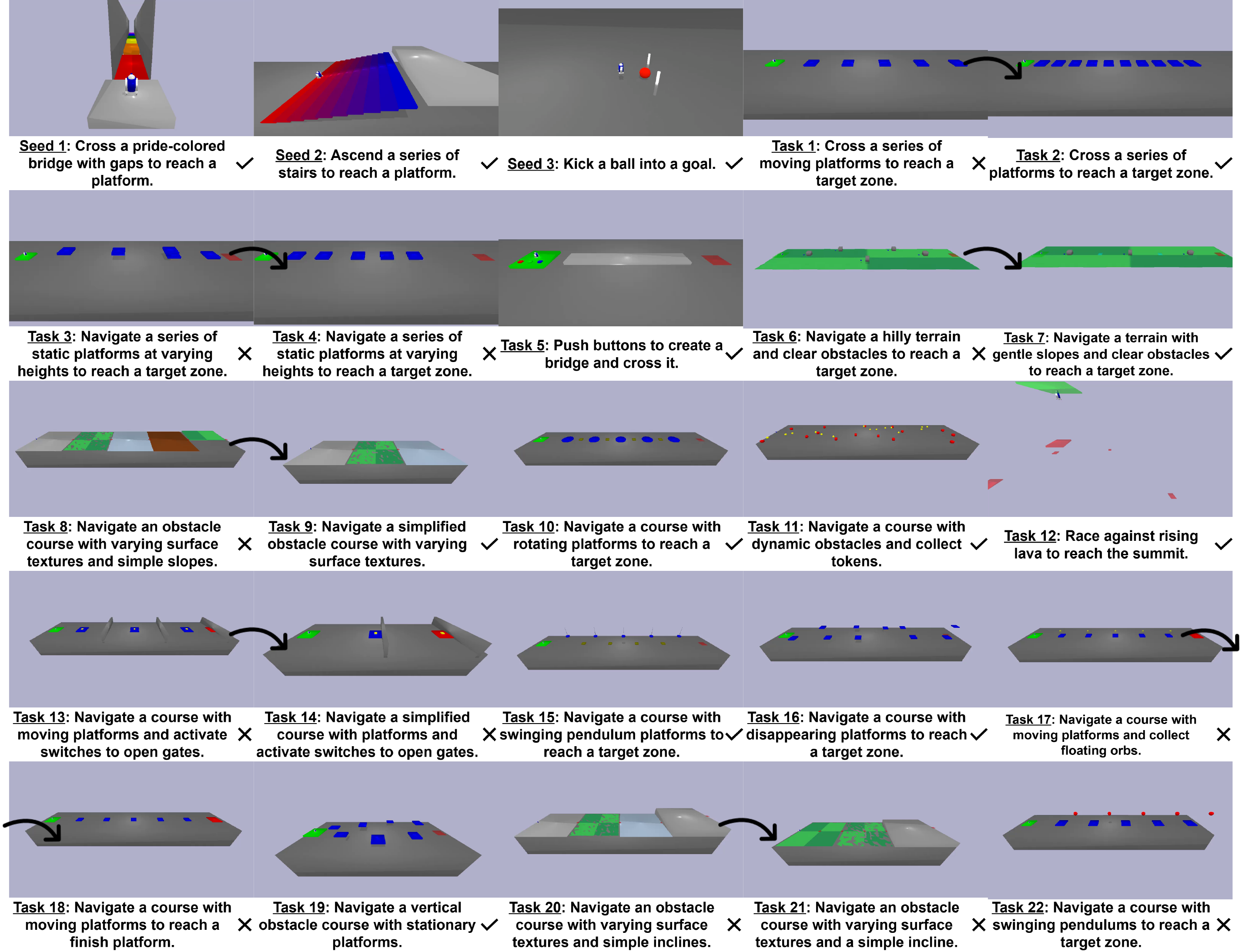}
    \caption{\textbf{Additional Short Run with Learning (Run 4).} This figure shows another instance of the short run experiment under the same settings as in \Cref{fig:results-short-run}. It demonstrates the progression of tasks generated by \ours{} and the learning outcomes of the RL agent. Checkmarks indicate successful learning, crosses indicate failures, and arrows show iterations on failed tasks.}
\end{figure}
\clearpage

\begin{figure}[h!]
    \centering
    \includegraphics[width=\textwidth]{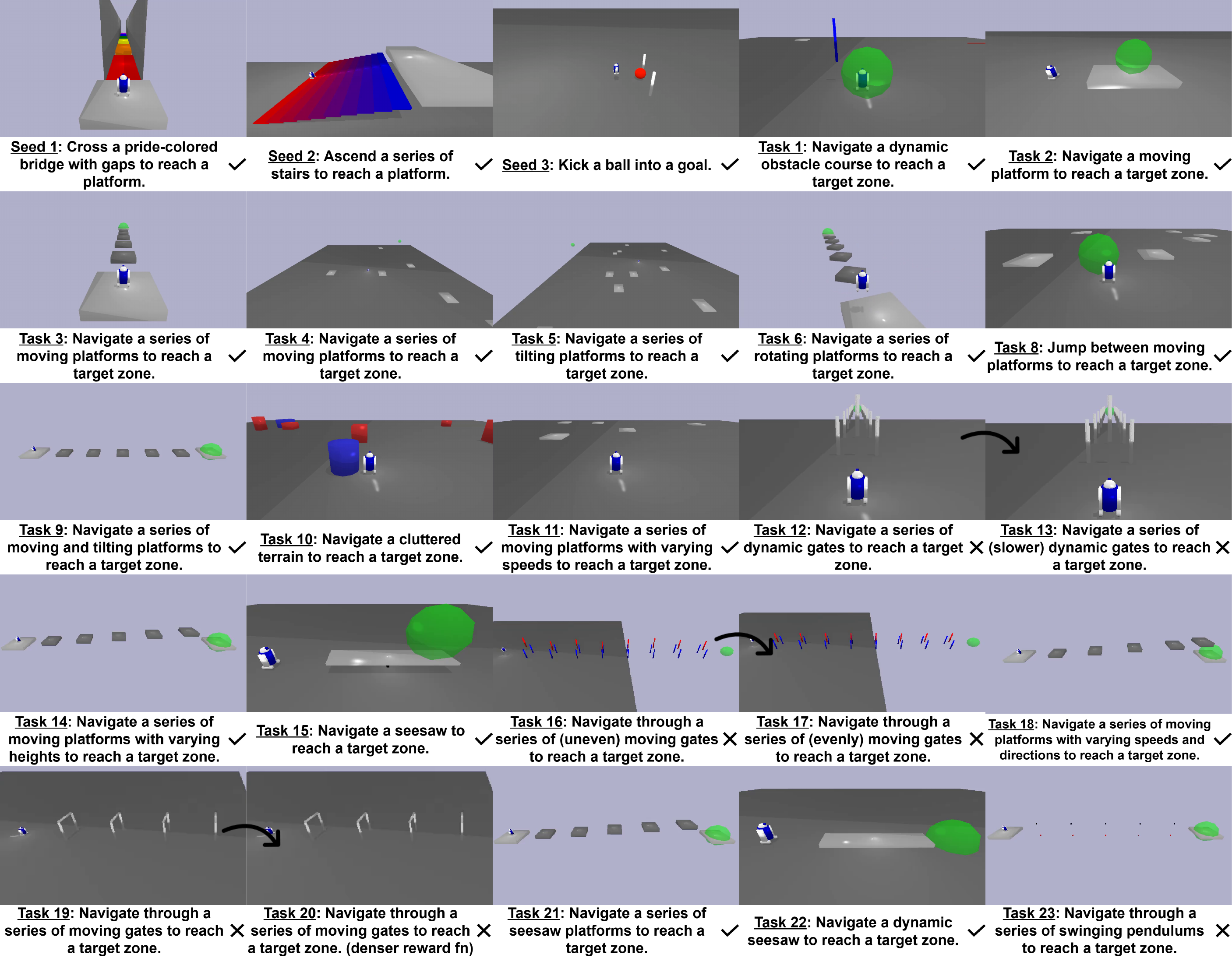}
    \caption{\textbf{Additional Short Run with Learning (Run 5).} This figure shows another instance of the short run experiment under the same settings as in \Cref{fig:results-short-run}. It demonstrates the progression of tasks generated by \ours{} and the learning outcomes of the RL agent. Checkmarks indicate successful learning, crosses indicate failures, and arrows show iterations on failed tasks.}
\end{figure}
\clearpage

\newpage
\section{Ablations on Short Runs with Learning}
\label{appendix:shortrun-ablation}

\subsection{Without Transfer Learning}

We conducted an experiment in which we ran OMNI-EPIC (i.e., with transfer learning between tasks) on 25 tasks with 5 replications and compared it to training policies from scratch (i.e., without transfer learning between tasks). We allocated the same total number of training steps (2 million) to the RL agent for each task. The only difference between the two settings was the policy initialization for each new task: in the transfer learning scenario, the RL policy was initialized from a checkpoint (a policy that had trained on another environment), whereas in the non-transfer learning scenario, it was initialized randomly (i.e., trained from scratch). OMNI-EPIC achieves higher success rate, with a median of 63.2\% (CI: 42.1 - 69.5) than training from scratch, with a median of 36.8\% (CI: 31.6 - 53.7). This demonstrates that the OMNI-EPIC agents build upon previously learned skills, creating a curriculum of increasing difficulty.

\subsection{Uniform Sampling Examples from the Archive}

\begin{figure}[h!]
    \centering
    \includegraphics[width=\textwidth]{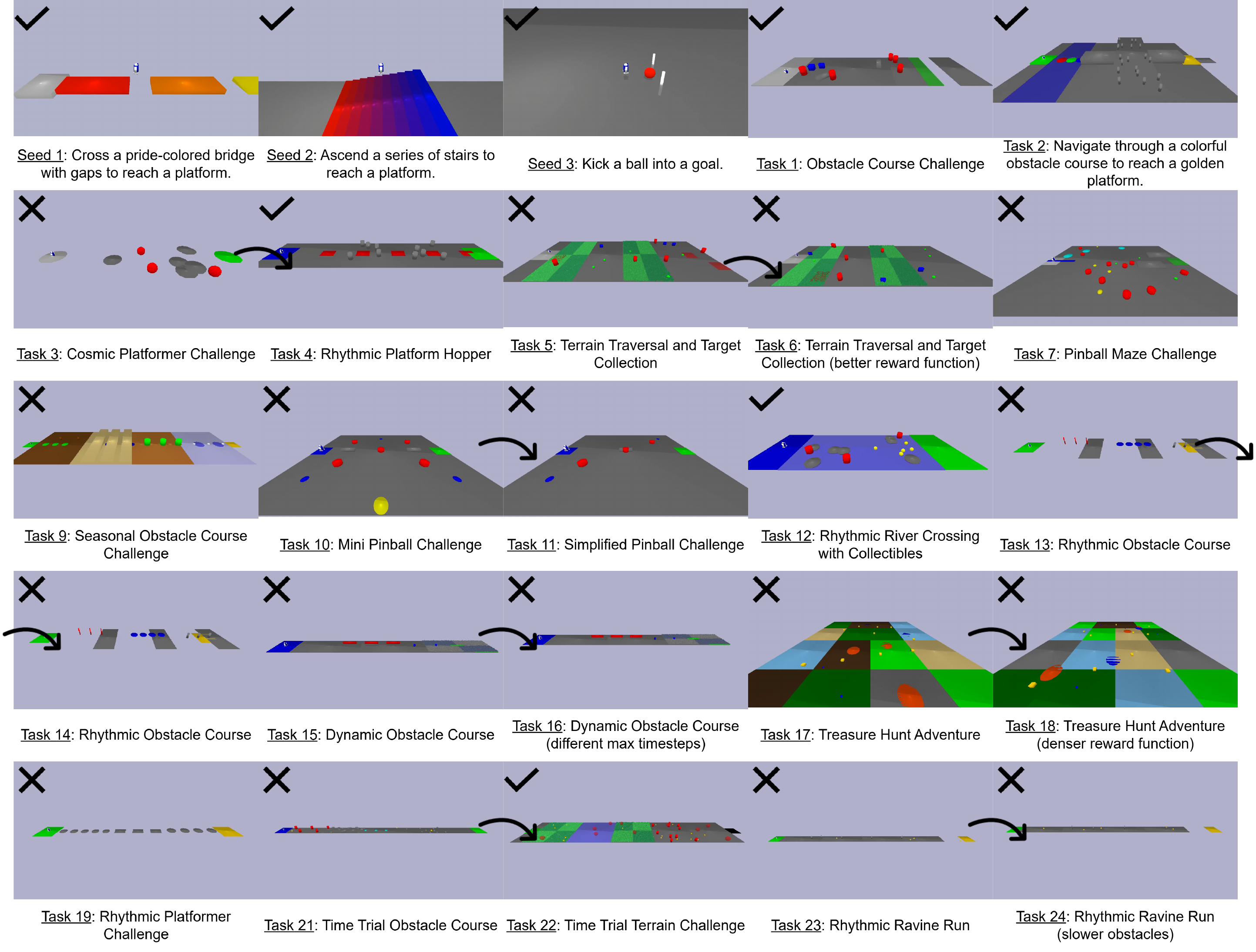}
    \caption{\textbf{Ablation of short run with learning, by sampling in-context examples uniformly.} This figure shows an ablation of the short run with learning experiment. The same settings as \Cref{fig:results-short-run} are used, except that, instead of using the most similar tasks as in-context examples for the task generator, we uniformly sample from the archive. Checkmarks indicate successful learning, crosses indicate failures, and arrows show iterations on failed tasks. Using uniform sampling for in-context examples results in learning only 5 of the generated tasks, much fewer than in \Cref{fig:results-short-run}, where the most similar tasks are used as in-context examples.}
\end{figure}
\clearpage

\subsection{Without Failed Examples}

\begin{figure}[h!]
    \centering
    \includegraphics[width=\textwidth]{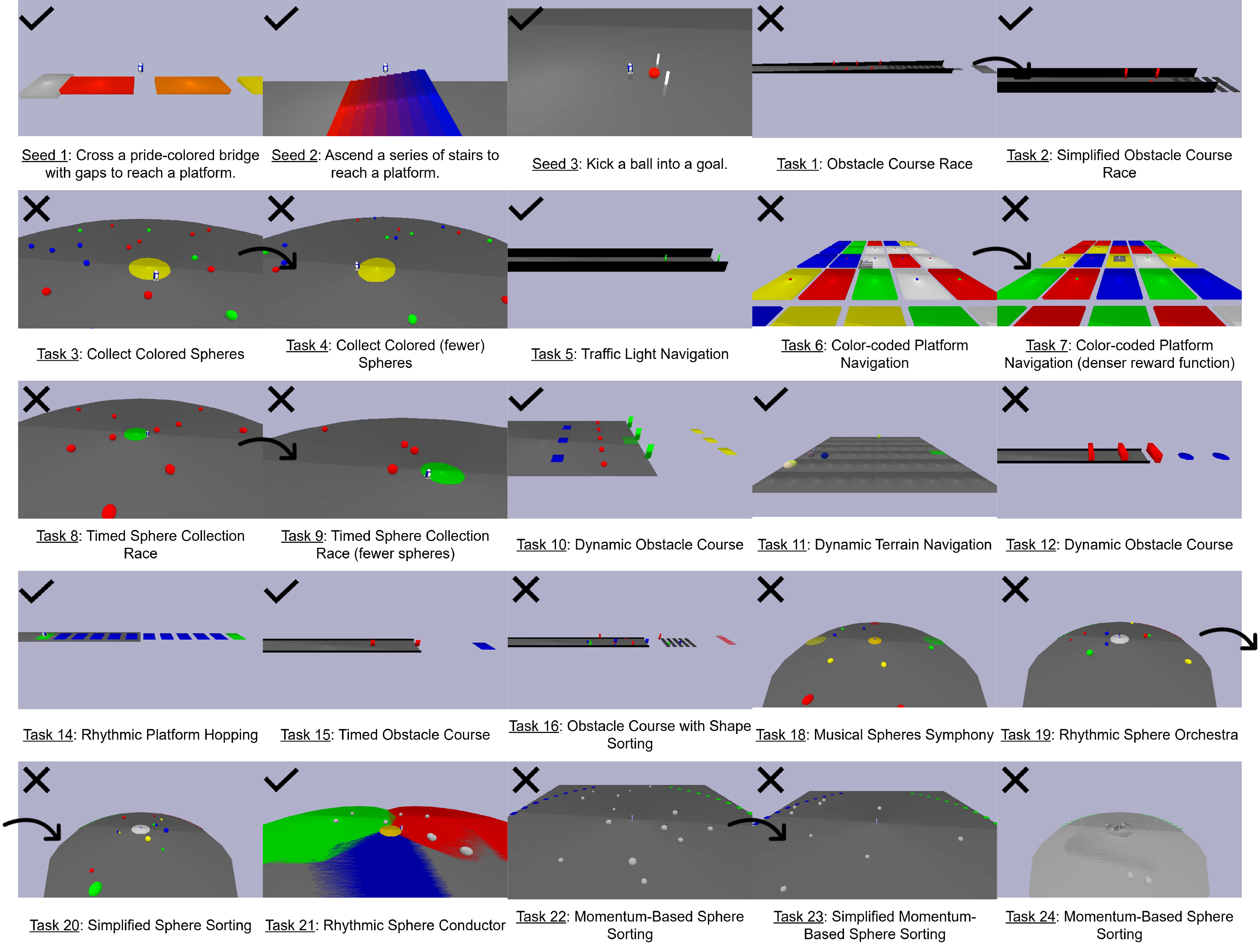}
    \caption{\textbf{Ablation of short run with learning, without failed examples input to the task generator.} This figure shows an ablation of the short run with learning experiment. The same settings as \Cref{fig:results-short-run} are used, except that no failed examples (only successful ones) are given to the task generator. Checkmarks indicate successful learning, crosses indicate failures, and arrows show iterations on failed tasks. We see that similar tasks are regenerated, even though the RL agent previously failed to learn them due to unsuitable reward function or incorrect environment configuration. Not giving the task generator examples of tasks that were attempted but failed results in learning only 7 of the generated tasks, much fewer than in \Cref{fig:results-short-run}, where both successful and failed examples are used as input.}
\end{figure}
\clearpage

\subsection{Without Learning Progress}

\begin{figure}[h!]
    \centering
    \includegraphics[width=\textwidth]{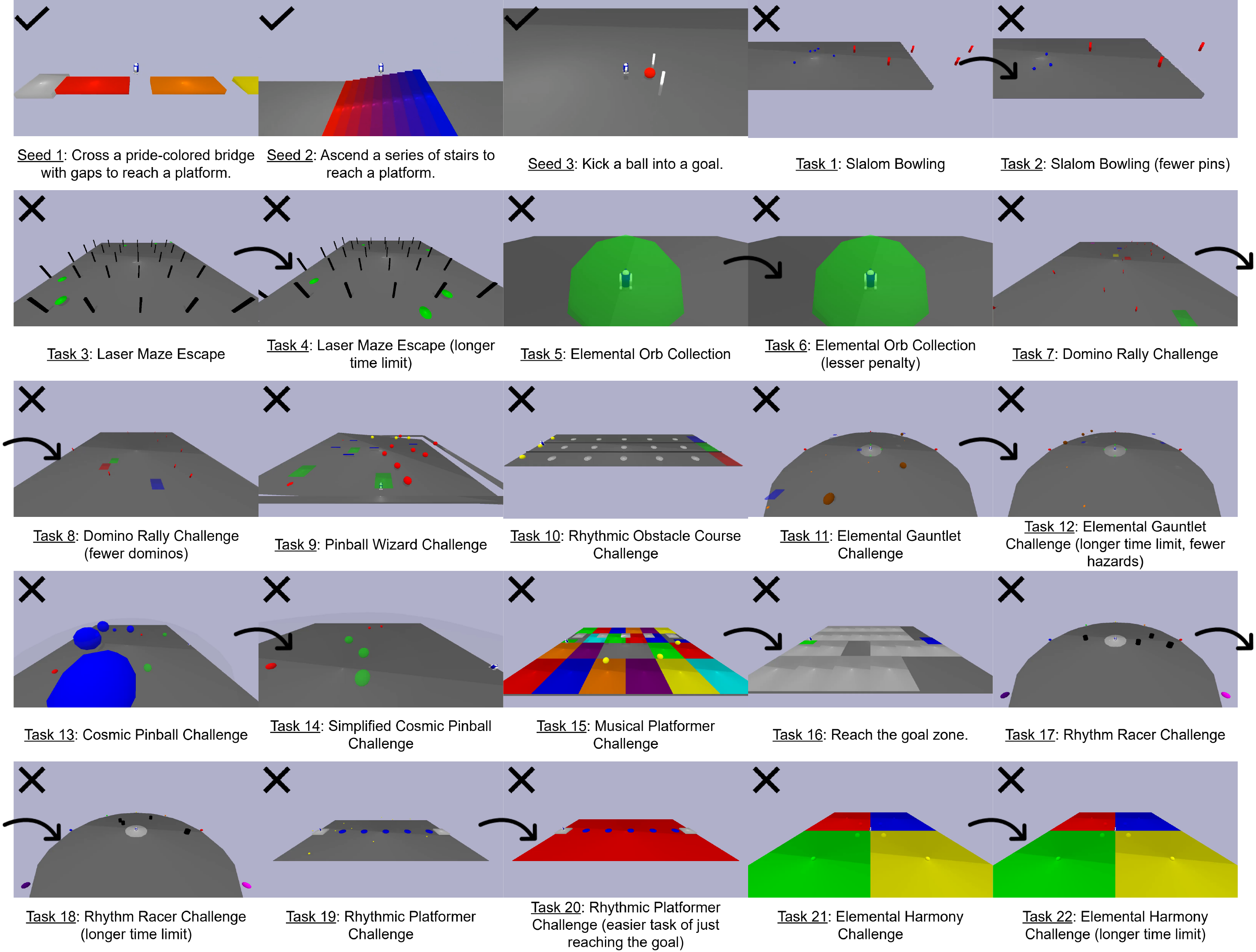}
    \caption{\textbf{Ablation of short run with learning, without learning progress notions in the task generator's prompt.} This figure shows an ablation of the short run with learning experiment. The same settings as \Cref{fig:results-short-run} are used, except that the task generator's prompt does not include any notion of learning progress. Checkmarks indicate successful learning, crosses indicate failures, and arrows show iterations on failed tasks. Without incorporating a notion of learning progress in the task generator's prompt, no tasks were successfully learned, which is much fewer than in \Cref{fig:results-short-run}, where the task generator's prompt includes both notions of interestingness and learning progress.}
\end{figure}
\textbf{System Prompt:}
\begin{lstlisting}[language={}]
You are an expert in reinforcement learning. Your goal is to help a robot master a diverse set of interesting tasks in simulation using PyBullet. You will be provided with the list of tasks that the robot has successfully learned, along with their corresponding environment code, and the list of tasks that the robot has attempted but failed to learn, along with their corresponding environment code. Your objective is to decide the next task for the robot, selecting one that is interesting and novel.

Instructions:
- The next task should be interesting:
    - Novel and creative compared to the tasks the robot has already tried.
    - Useful according to humans.
    - Design rich environments with a large number of diverse objects and terrains, and with a clear task for the robot to execute.
    - The task should be fun or engaging to watch. You can draw inspiration from real-world tasks or video games. Be creative!
- Be specific in the task description:
    - State clearly what the task of the robot is.
    - Define clearly what the success condition is.
    - Define clearly what are the different reward and penalty components.
    - Define clearly what the termination conditions are. If the reward components include a survival reward, ensure the episode only terminates when the agent fails the task.
- The task should not take too long to complete.
- The robot can push objects around but lacks the ability to grab, pick up, carry, or stack objects. Don't suggest tasks that involve these skills.
- Don't suggest tasks that require the robot to navigate through a maze.
- If the task involves navigating a terrain with obstacles, make sure that the robot can not go around the obstacles.
- If the task involves a target zone, make sure that the collision of the target zone is set to False.
- Return only the task description, not the environment code.
- Ensure that the designs pose no harm to humans and align with human values and ethics.

Robot description:
{ROBOT_DESC}

Desired format:
Reasoning for what the next task should be:
<reasoning>

Next task description:
\"\"\"
<task description>
\"\"\"
\end{lstlisting}

\newpage
\section{Human Evaluation Setup}
\label{appendix:humaneval-succdet}

To evaluate the alignment between human judgment and the automated success detector, we conducted a study with 50 participants. The goal was to assess whether human evaluators agreed with the success detector’s assessments of task completion by the robot. Each participant reviewed videos of a robot attempting various tasks, alongside the corresponding task descriptions. Below are the detailed instructions and setup used for the evaluation.

\begin{figure}[h!]
    \centering
    \includegraphics[width=0.45\textwidth]{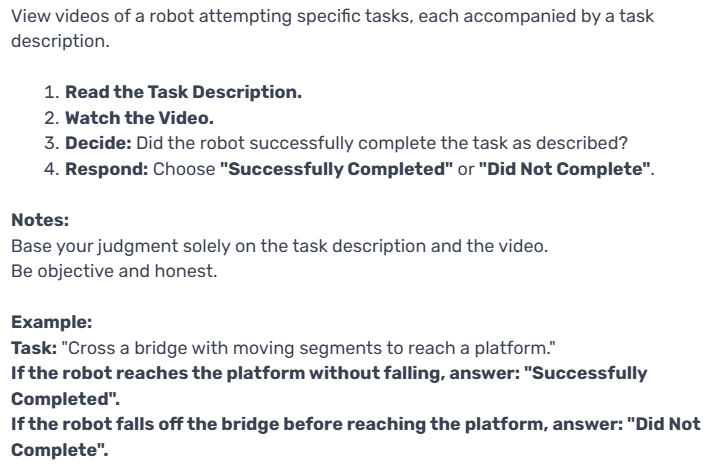}
    \includegraphics[width=0.45\textwidth]{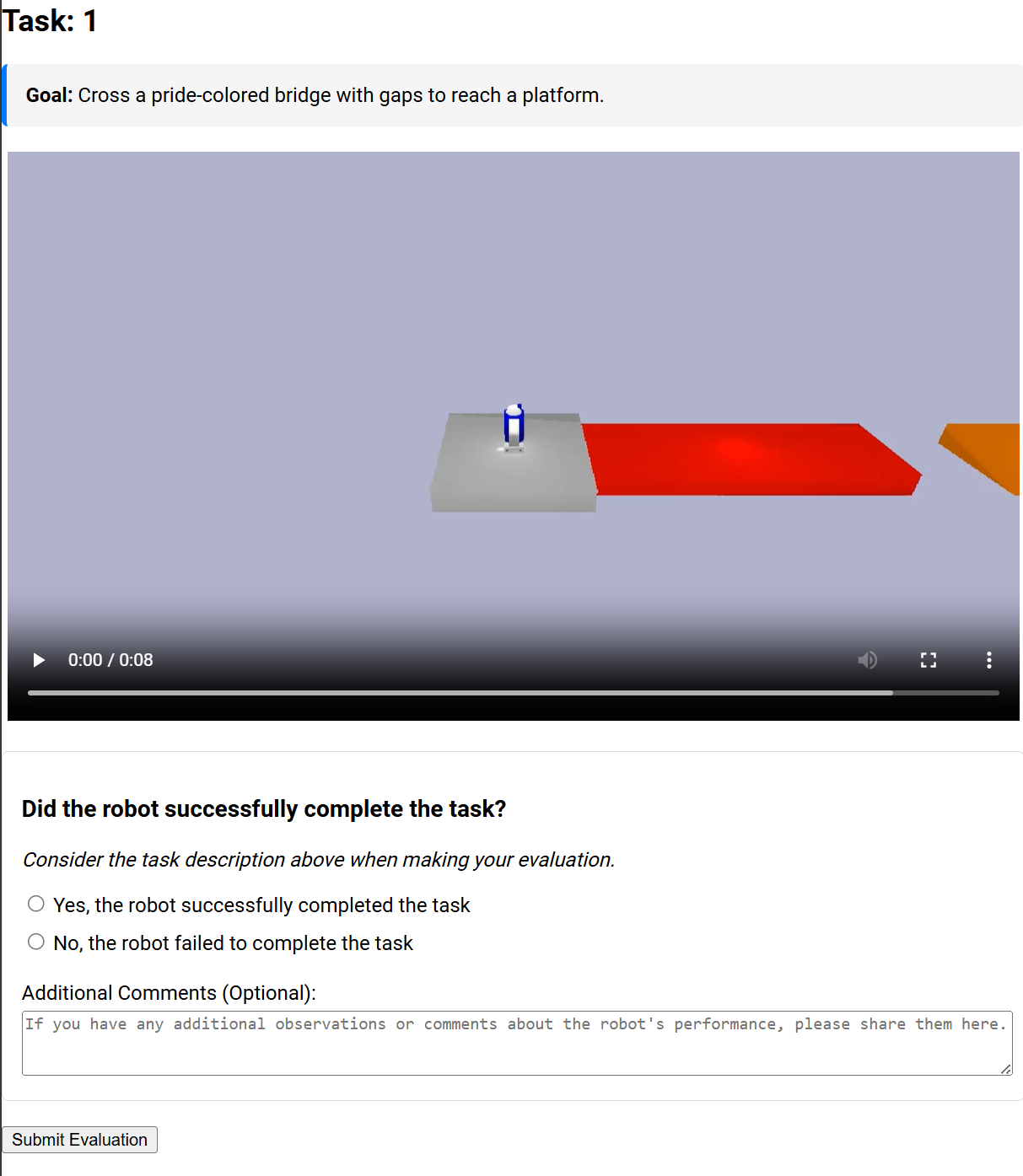}
    \caption{\textbf{Human evaluation setup.} (Left) Instructions given to participants. (Right) Annotation interface used by participants to annotate the data.}
\end{figure}

The human evaluation study was conducted using the \texttt{cloudresearch.com} platform. Participants were compensated \$0.25 for each response (estimated hourly rate of \$15.00/hour, as each response took less than 1 minute), ensuring fair and ethical payment for their time and effort. The study aimed to collect accurate and unbiased assessments to compare with the success detector's automated evaluations.

\newpage
\section{Supplementary Materials for Quantitative Results}
\label{appendix:quantitative}

\begin{figure}[h!]
    \centering
    \includegraphics[width=\textwidth]{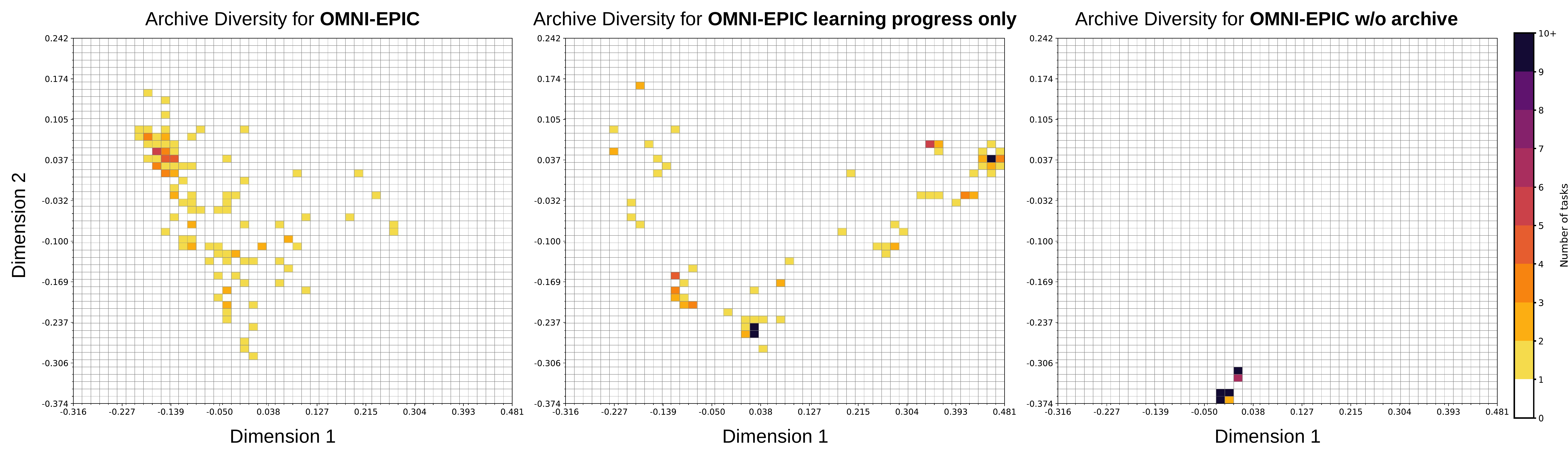}
    \caption{\textbf{Archive diversity results.} Archive diversity plots of long runs with simulated learning of OMNI-EPIC and the controls. Fewer tasks fall into the same discretized cells for OMNI-EPIC than OMNI-EPIC Learning Progress only or OMNI-EPIC w/o archive. The substantial difference between the left and center plots is more easily observed in \Cref{fig:quantitative}.}
\end{figure}

\begin{figure}[h!]
    \centering
    \includegraphics[width=\textwidth]{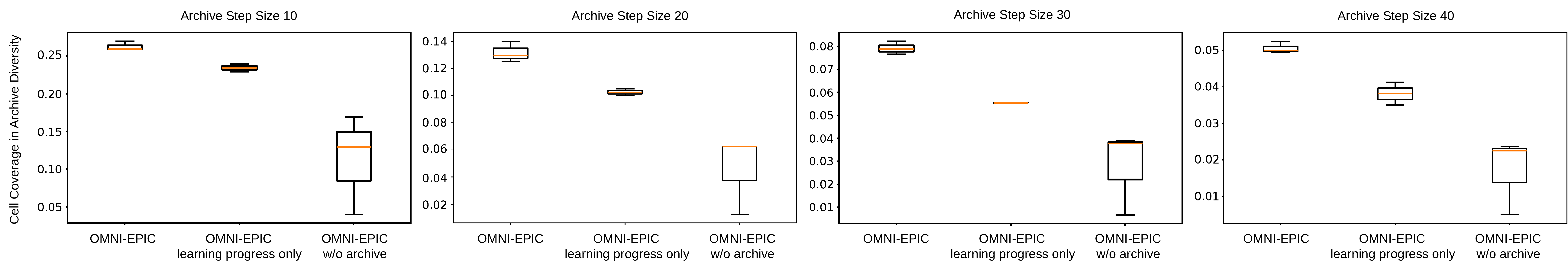}
    \caption{\textbf{Cell coverage of archive diversity plots with different discretization levels for long runs with simulated learning.} This figure is similar to \Cref{fig:quantitative}, which uses an archive discretization level of 50, but here we present cell coverage results for archive diversity plots with discretization levels of [10, 20, 30, 40] across methods on long runs with simulated learning. OMNI-EPIC consistently achieves significantly higher cell coverage compared to the controls, even across different archive discretization levels (p < 0.05, Mann-Whitney U test).}
    \label{fig:cell_coverage_stepsizes}
\end{figure}

\begin{figure}[h!]
    \centering
    \includegraphics[width=0.8\textwidth]{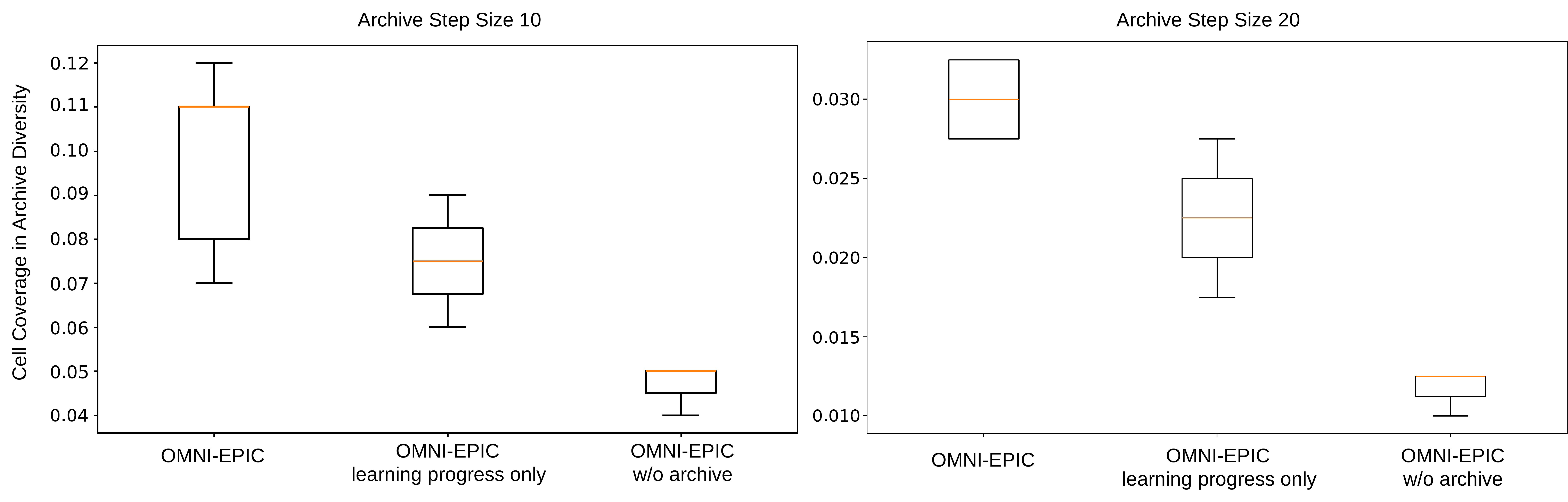}
    \caption{\textbf{Cell coverage of archive diversity plots with different discretization levels for short runs with learning.} This figure is similar to \Cref{fig:quantitative}, which uses an archive discretization level of 50 for long run with simulated learning, but here we present cell coverage results for archive diversity plots with discretization levels of [10, 20] across methods on short runs with learning. While we see similar trends as \Cref{fig:quantitative} and \Cref{fig:cell_coverage_stepsizes}, the differences between methods are not always statistically significant (not all p < 0.05, Mann-Whitney U test). This is due to the shorter training runs, as the effects of OMNI-EPIC become more pronounced over longer runs.}
\end{figure}

\begin{figure}[h!]
    \centering
    \includegraphics[width=\textwidth]{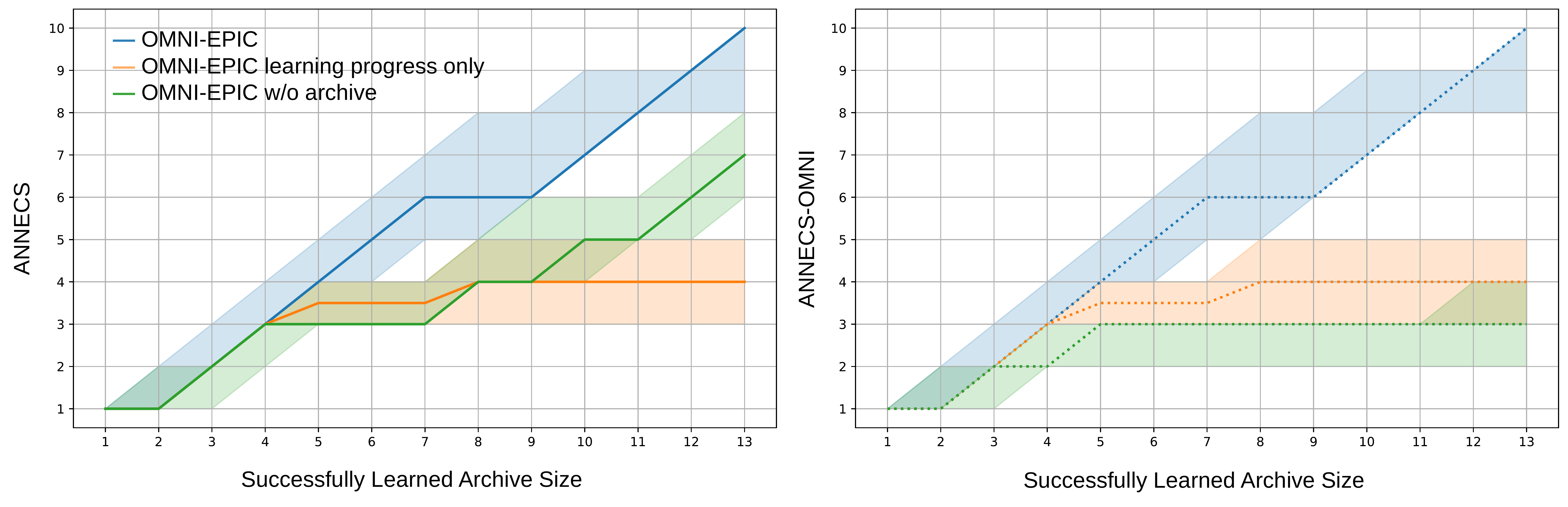}
    \caption{\textbf{(Left) ANNECS and (Right) ANNECS-OMNI results.} Short runs with RL training by OMNI-EPIC and the controls are repeated five times. Darker lines are median values, shaded regions are 95\% confidence intervals. OMNI-EPIC significantly outperforms the controls on both metrics. There is no difference between ANNECS and ANNECS-OMNI for OMNI-EPIC, indicating that all tasks learned by OMNI-EPIC are considered interesting.}
\end{figure}

\begin{figure}[h!]
    \centering
    \includegraphics[width=0.7\textwidth]{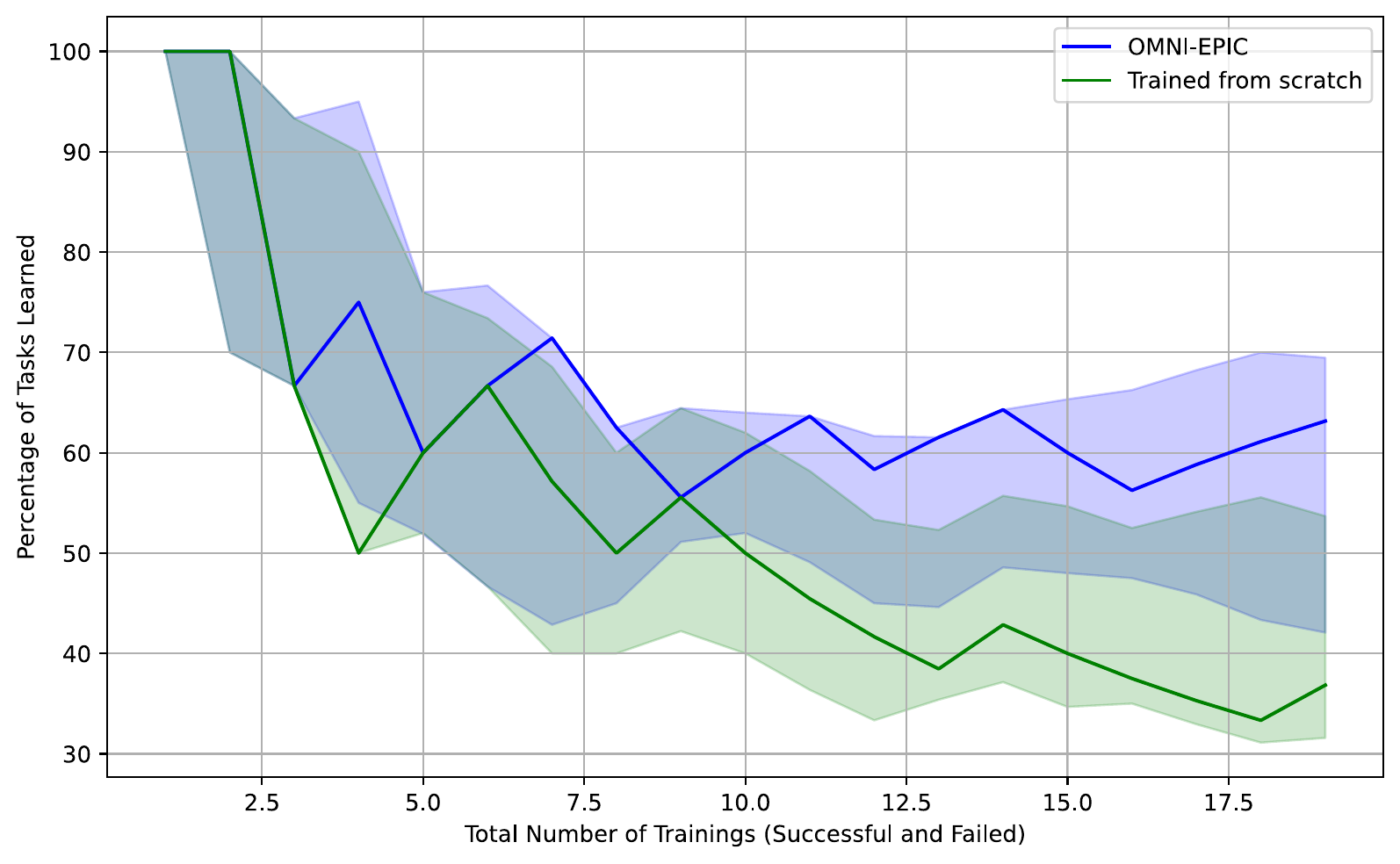}
    \caption{\textbf{Percentage of tasks learned over attempted tasks in short runs with learning experiments.} Attempted tasks include all tasks that were attempted with RL training. Short runs with RL training by OMNI-EPIC are repeated five times. The blue line represents \ours{}, while the green line represents using the same generated tasks from \ours{} and training each task from scratch. The darker lines indicate median values, and the shaded regions represent 95\% confidence intervals. The decreasing percentage of learned tasks, along with the widening gap between tasks learned by OMNI-EPIC and those trained from scratch, shows that \ours{} generates increasingly difficult tasks over time.}
\end{figure}
\clearpage

\newpage
\section{Selecting the most similar tasks}
\label{appendix:selecting-similar-tasks}

To generate new tasks that are both relevant and challenging, it is essential to select tasks from the archive as a reference. This selection process ensures that the generated tasks build upon the agent's prior knowledge and skills. \ours{}'s process of selecting the most similar tasks, given a query task, takes inspiration from \citet{lewis2020retrieval}. The first step is to embed all tasks (both their natural language descriptions and environment code) within the archive into a high-dimensional vector space using a pretrained language encoder. This embedding allows us to represent each task as a dense vector that captures its semantic meaning and characteristics. Once all tasks are embedded, we can efficiently compare the similarity between tasks using cosine similarity. For a given query task, we retrieve a predefined number of tasks from the archive that exhibit the closest cosine similarity to the query's embedding, ensuring that the selected tasks are the most relevant and similar to the current learning context.

A potential limitation of our approach is the possibility of cyclic behavior where the task generator alternates between generating task types that are consistently rejected due to their similarity to existing tasks in the archive. While this scenario is unlikely in practice due to the stochastic nature of task generation and the diversity of the environment distribution, it cannot be completely ruled out. Furthermore, as context lengths increase with advancements in FM capabilities, the likelihood of such cyclic behavior would diminish even more.

\newpage
\section{Hyperparameters}
\label{appendix:hyperparameters}

\begin{table}[h]
\label{tab:hyperparameters-ours}
\caption{\ours{} hyperparameters}
\centering
\begin{tabular}{l l c}
\toprule
\textbf{Component} & \textbf{Parameter} & \textbf{Value}\\
\midrule
\multirow{2}{*}{General} & Max number of iterations & $5$ \\
 & Embedding method & {\begin{tabular}[c]{@{}c@{}}OpenAI\\(text-embedding-3-small)\end{tabular}}\\
\midrule
\multirow{5}{*}{Task Generator} & Number of learned examples & $5$\\
 & Number of failed examples & $5$\\
 & Client & Anthropic\\
 & Model & {\begin{tabular}[c]{@{}c@{}}Claude 3 Opus\\(claude-3-opus-20240229)\end{tabular}}\\
 & Temperature & 0\\
\midrule
\multirow{4}{*}{Environment Generator} & Number of few-shot examples & $5$\\
 & Client & Anthropic\\
 & Model & {\begin{tabular}[c]{@{}c@{}}Claude 3 Opus\\(claude-3-opus-20240229)\end{tabular}}\\
 & Temperature & 0\\
\midrule
\multirow{4}{*}{Post-Generation MoI} & Number of similar tasks & $10$\\
 & Client & OpenAI\\
 & Model & {\begin{tabular}[c]{@{}c@{}}GPT-4o\\(gpt-4o-2024-05-13)\end{tabular}}\\
 & Temperature & $0$\\
\midrule
\multirow{3}{*}{Success Detector} & Client & OpenAI\\
 & Model & {\begin{tabular}[c]{@{}c@{}}GPT-4o\\(gpt-4o-2024-05-13)\end{tabular}}\\
 & Temperature & $0$\\
\midrule
\multirow{5}{*}{Task Reflection} & Max number of iterations & $1$\\
 & Number of few-shot examples & $5$\\
 & Client & OpenAI\\
 & Model & {\begin{tabular}[c]{@{}c@{}}GPT-4o\\(gpt-4o-2024-05-13)\end{tabular}}\\
 & Temperature & $0$\\
\bottomrule
\end{tabular}
\end{table}

\begin{table}[h]
\label{tab:hyperparameters-dreamer}
\caption{DreamerV3 hyperparameters}
\centering
\begin{tabular}{l c}
\toprule
\textbf{Parameter} & \textbf{Value}\\
\midrule
Total time steps & $2 \times 10^{6}$ \\
Replay buffer size & $10^6$\\
Batch size & $16$\\
Batch length & $64$\\
Discount factor & $0.997$\\
Learning rate & $3 \times 10^{-4}$\\
\bottomrule
\end{tabular}
\end{table}

\textbf{Compute Resources}\\
For each task, the R2D2 agent is trained for approximately 1 hour using 2 NVIDIA RTX 6000 Ada Generation GPUs and 32 CPU cores.

\newpage
\section{Prompts}
\label{appendix:prompts}
\subsection{Task Generator}
\label{appendix:task-gen-prompts}
\textbf{System Prompt:}
\begin{lstlisting}[language={}]
You are an expert in curriculum learning and reinforcement learning. Your goal is to help a robot master a diverse set of interesting tasks in simulation using PyBullet. You will be provided with the list of tasks that the robot has successfully learned, along with their corresponding environment code, and the list of tasks that the robot has attempted but failed to learn, along with their corresponding environment code. Your objective is to decide the next task for the robot, selecting one that will maximize learning effectiveness based on its past successes and failures.

Instructions:
- The next task should be learnable:
    - Not too difficult for the robot to learn given its current skill set.
    - Realistic for the robot based on its description.
    - Possible to complete in simulation in PyBullet.
- The next task should be interesting, i.e., either:
    - Novel compared to the tasks the robot has already learned. You can either add complexity gradually on an existing task or design a radically novel task from scratch.
    - Useful according to humans, making it worth learning.
    - Creative or surprising.
    - Optionally, the task can be fun and engaging to watch.
- Be specific in the task description:
    - State clearly what the task of the robot is.
    - If the task involves objects, be specific about their positions and orientations relative to the robot. Be careful to avoid collisions between objects or with the robot when you decide on the initial positions.
    - If the task involves dynamically moving objects, be specific about their movement.
- You can draw inspiration from real-world tasks or video games. Be creative!
- The task should not take too long to complete.
- The robot can push objects around but lacks the ability to grab, pick up, carry, or stack objects. Don't suggest tasks that involve these skills.
- Don't suggest tasks that require the robot to navigate through a maze.
- Return only the task description, not the environment code.
- Ensure that the designs pose no harm to humans and align with human values and ethics.

Robot description:
{ROBOT_DESC}

Desired format:
Reasoning for what the next task should be:
<reasoning>

Next task description:
"""
<task description>
"""
\end{lstlisting}

\textbf{User Prompt:}
\begin{lstlisting}[language={}]
Environment code examples:
{ENV_CODES_EXAMPLE}

Learned tasks and environment code:
{ENV_CODES_LEARNED}

Failed tasks and environment code:
{ENV_CODES_FAILED}
\end{lstlisting}

\subsection{Environment Generator}
\label{appendix:env-gen-prompts}
\textbf{System Prompt:}
\begin{lstlisting}[language={}]
You are an expert in Python programming and reinforcement learning. Your goal is to implement an environment in PyBullet specifically designed to train a robot for a given task. You will be provided with the task description and with pairs of task description and environment code. Your objective is to write environment code that rigorously aligns with the task description, helping the robot learn the task as effectively as possible.

Instructions:
- Write code without using placeholders.
- Don't change the import statements.
- For each object, always define its size first, and ensure the object's initial position is set relative to the platform it starts on or any other object, as demonstrated in the provided environment code examples. For example, if an object is initialized on the ground, define its position as: [self.platform_position[0], self.platform_position[1], self.platform_position[2] + self.platform_size[2] / 2 + self.object_size[2] / 2].
- Ensure the robot's initial position is set relative to the platform it starts on, as demonstrated in the provided environment code examples. For example, if the robot starts on a platform, its initial position should be set to [self.platform_position[0], self.platform_position[1], self.platform_position[2] + self.platform_size[2] / 2 + self.robot.links["base"].position_init[2]].
- If the task involves navigating a terrain with obstacles, make sure that the robot cannot go around the obstacles.
- Implement the methods `Env.reset()`, `Env.step()`, `Env.get_task_rewards()`, `Env.get_terminated()`, `Env.get_success()`. You can implement additional methods if needed.
- `Env.get_task_rewards()` returns a dictionary with the different reward components to help the robot learn the task. You should implement dense reward components that are easy to optimize and defined in the range -10. to 10.
- `Env.get_terminated()` returns a boolean that indicates whether the episode is terminated.
- `Env.get_success()` returns a boolean that indicates whether the task is successfully completed.

Robot description:
{ROBOT_DESC}

Desired format:
Environment code:
```python
<environment code>
```
\end{lstlisting}

\textbf{User Prompt:}
\begin{lstlisting}[language={}]
Pairs of task description and environment code:
{ENV_CODES_EXAMPLE}

Task description:
{TASK_DESC}
\end{lstlisting}

\subsection{Environment Generator Reflection}
\label{appendix:env-gen-reflect-prompts}
\textbf{System Prompt:}
\begin{lstlisting}[language={}]
You are an expert in Python programming and reinforcement learning. Your goal is to implement an environment in PyBullet specifically designed to train a robot for a given task. You will be provided with environment code examples, with an environment code that returns an error when executed and with the specific error that was encountered. Your objective is to reason about the error and provide a new, improved environment code with no error.

Instructions:
- Write code without using placeholders.
- Don't change the import stateme nts.
- For each object, always define its size first, and ensure the object's initial position is set relative to the platform it starts on or any other object, as demonstrated in the provided environment code examples. For example, if an object is initialized on the ground, define its position as: [self.platform_position[0], self.platform_position[1], self.platform_position[2] + self.platform_size[2] / 2 + self.object_size[2] / 2].
- Ensure the robot's initial position is set relative to the platform it starts on, as demonstrated in the provided environment code examples. For example, if the robot starts on a platform, its initial position should be set to [self.platform_position[0], self.platform_position[1], self.platform_position[2] + self.platform_size[2] / 2 + self.robot.links["base"].position_init[2]].
- If the task involves navigating a terrain with obstacles, make sure that the robot cannot go around the obstacles.
- Implement the methods `Env.reset()`, `Env.step()`, `Env.get_task_rewards()`, `Env.get_terminated()`, `Env.get_success()`. You can implement additional methods if needed.
- `Env.get_task_rewards()` returns a dictionary with the different reward components to help the robot learn the task. You should implement dense reward components that are easy to optimize and defined in the range -10. to 10.
- `Env.get_terminated()` returns a boolean that indicates whether the episode is terminated.
- `Env.get_success()` returns a boolean that indicates whether the task is successfully completed.

Robot description:
{ROBOT_DESC}

Desired format:
How to solve the error:
<reasoning>

Environment code:
```python
<environment code>
```
\end{lstlisting}

\textbf{User Prompt:}
\begin{lstlisting}[language={}]
Environment code examples:
{ENV_CODES_EXAMPLE}

Environment code with error:
{ENV_CODE}

Error:
{ERROR}
\end{lstlisting}

\subsection{Post-Generation Model of Interestingness}
\label{appendix:model-of-int-prompts}
\textbf{System Prompt:}
\begin{lstlisting}[language={}]
You are an expert in curriculum learning and reinforcement learning. Your goal is to help a robot master a diverse set of interesting tasks in simulation using PyBullet. You will be provided with a list of old tasks and with a new task. Your objective is to determine whether the new task is interesting or not.

The new task can be considered interesting if one of the following is true, the new task is:
- Novel compared to the old tasks, to build a diverse skill set.
- Creative or surprising.
- Fun or engaging to watch.
- Not too easy for the robot to learn given its current skill set, progressing toward more complex challenges.
- Useful according to humans, making it worth learning.

Robot description:
{ROBOT_DESC}

Desired format:
Reasoning for why the new task is interesting or not:
<reasoning>

Is the new task interesting?:
<Yes/No>
\end{lstlisting}

\textbf{User Prompt:}
\begin{lstlisting}[language={}]
Old tasks:
{ENV_CODES_EXAMPLE}

New task:
{ENV_CODE}
\end{lstlisting}

\newpage
\section{Few-shot Examples}
\label{appendix:few-shot-examples}
\begin{lstlisting}[language=Python]
import numpy as np
from oped.envs.r2d2.base import R2D2Env


class Env(R2D2Env):
    """
    Cross a pride-colored bridge to reach a platform.

    Description:
    - A start platform and an end platform (each 3 m in size and 0.5 m in thickness) are placed 30 m apart.
    - The two platforms are connected by a bridge (2 m wide) divided in multiple segments. Each segment has a different color corresponding to the pride colors.
    The robot is initialized on the start platform.
    The task of the robot is to cross the bridge to reach the end platform as fast as possible.

    Success:
    The task is successfully completed when the robot reaches the end platform.

    Rewards:
    To help the robot complete the task:
    - The robot receives a reward for each time step it remains on the bridge or platforms, encouraging steady progress.
    - The robot is rewarded based on how much it reduces the distance to the end platform, incentivizing swift movement towards the goal.

    Termination:
    The task terminates immediately if the robot falls off the start platform, any segment of the bridge, or the end platform.
    """

    def __init__(self):
        super().__init__()

        # Init start platform
        self.platform_size = [3., 3., 0.5]
        self.platform_start_position = [0., 0., 0.]
        self.platform_end_position = [self.platform_start_position[0] + 30., self.platform_start_position[1], self.platform_start_position[2]]
        self.platform_start_id = self.create_box(mass=0., half_extents=[self.platform_size[0] / 2, self.platform_size[1] / 2, self.platform_size[2] / 2], position=self.platform_start_position, color=[0.8, 0.8, 0.8, 1.])
        self.platform_end_id = self.create_box(mass=0., half_extents=[self.platform_size[0] / 2, self.platform_size[1] / 2, self.platform_size[2] / 2], position=self.platform_end_position, color=[0.8, 0.8, 0.8, 1.])

        # Init bridge
        self.bridge_length = self.platform_end_position[0] - self.platform_start_position[0] - self.platform_size[0]
        self.bridge_width = 2.
        pride_colors = [
            [1.0, 0.0, 0.0, 1.],  # Red
            [1.0, 0.5, 0.0, 1.],  # Orange
            [1.0, 1.0, 0.0, 1.],  # Yellow
            [0.0, 0.5, 0.0, 1.],  # Green
            [0.0, 0.0, 1.0, 1.],  # Blue
            [0.7, 0.0, 1.0, 1.],  # Violet
        ]

        # Segment length
        num_colors = len(pride_colors)
        segment_size = self.bridge_length / num_colors

        # Create segments
        for i, color in enumerate(pride_colors):
            segment_id = self.create_box(mass=0., half_extents=[segment_size / 2, self.bridge_width / 2, self.platform_size[2] / 2], position=[self.platform_start_position[0] + self.platform_size[0] / 2 + segment_size / 2 + i * segment_size, self.platform_start_position[1], self.platform_start_position[2]], color=color)
            self._p.changeDynamics(bodyUniqueId=segment_id, linkIndex=-1, lateralFriction=0.8, restitution=0.5)

    def create_box(self, mass, half_extents, position, color):
        collision_shape_id = self._p.createCollisionShape(shapeType=self._p.GEOM_BOX, halfExtents=half_extents)
        visual_shape_id = self._p.createVisualShape(shapeType=self._p.GEOM_BOX, halfExtents=half_extents, rgbaColor=color)
        return self._p.createMultiBody(baseMass=mass, baseCollisionShapeIndex=collision_shape_id, baseVisualShapeIndex=visual_shape_id, basePosition=position)

    def get_object_position(self, object_id):
        return np.asarray(self._p.getBasePositionAndOrientation(object_id)[0])

    def get_distance_to_object(self, object_id):
        object_position = self.get_object_position(object_id)
        robot_position = self.robot.links["base"].position
        return np.linalg.norm(object_position[:2] - robot_position[:2])

    def reset(self):
        observation = super().reset()

        # Reset robot position on start platform
        self._p.resetBasePositionAndOrientation(self.robot.robot_id, [self.platform_start_position[0], self.platform_start_position[1], self.platform_start_position[2] + self.platform_size[2] / 2 + self.robot.links["base"].position_init[2]], self.robot.links["base"].orientation_init)

        return observation

    def step(self, action):
        # Before taking action
        self.distance_to_platform_end = self.get_distance_to_object(self.platform_end_id)

        observation, reward, terminated, truncated, info = super().step(action)

        return observation, reward, terminated, truncated, info

    def get_task_rewards(self, action):
        # After taking action
        new_distance_to_platform_end = self.get_distance_to_object(self.platform_end_id)

        # Survival
        survival = 1.

        # Reach end platform
        reach_platform_end = (self.distance_to_platform_end - new_distance_to_platform_end) / self.dt

        return {"survival": survival, "reach_platform_end": reach_platform_end}

    def get_terminated(self, action):
        # Terminate if fall off
        return self.robot.links["base"].position[2] < self.platform_start_position[2]

    def get_success(self):
        # Success if reach end platform
        is_on_platform_end = self.get_distance_to_object(self.platform_end_id) < self.platform_size[2] / 2
        return is_on_platform_end
\end{lstlisting}

\begin{lstlisting}[language=Python]
import numpy as np
from oped.envs.r2d2.base import R2D2Env


class Env(R2D2Env):
    """
    Cross over lava on a boat to reach a target zone.

    Description:
    - The lava is simulated with an orange, 10 x 10 m heightfield.
    - There are two platforms on either side of the lava, each measuring 5 x 10 m. One serves as the start platform and the other as the end platform.
    - The boat is a box with dimensions 3 meters in length, 2 meters in width, and 0.2 meters in height. It is initialized next to the start platform at a random y-position.
    - The boat has a button that, when pressed, activates the boat to move over the lava at a speed of 3 meters per second.
    - The end platform has a target zone indicated by a green, transparent sphere.
    The robot's task is to jump onto the boat from the start platform, press the button to activate the boat, and travel across the lava to reach the end platform. The robot must then enter the target zone to complete the task.

    Success:
    The task is successfully completed when the robot enters the target zone on the end platform.

    Rewards:
    To guide the robot to complete the task:
    - The robot receives a reward for each time step it remains active and does not fall off or touch the lava.
    - The robot is rewarded for making progress towards pressing the button on the boat.
    - Additional rewards are given for progressing towards the target zone, with a significant bonus for entering the target zone.

    Termination:
    The task terminates immediately if the robot falls off the platform or the boat, or if it touches the simulated lava.
    """

    def __init__(self):
        super().__init__()

        # Init lava
        self.lava_size = [10., 10.]
        self.lava_height = 0.1
        self.lava_position = [0., 0., 0.]
        self.lava_id = self.create_heightfield(
            size=self.lava_size,
            height_max=self.lava_height,  # create small bumps to create a fluid-like surface
            position=self.lava_position,
            resolution=20,  # number of points per meter
            repeats=2,
        )
        self._p.changeVisualShape(objectUniqueId=self.lava_id, linkIndex=-1, rgbaColor=[1., 0.3, 0.1, 1.])  # change to lava color

        # Init platforms
        self.platform_size = [5., self.lava_size[1], 1.]
        self.platform_start_position = [self.lava_position[0] - self.lava_size[0] / 2 - self.platform_size[0] / 2, self.lava_position[1], self.lava_position[2]]
        self.platform_end_position = [self.lava_position[0] + self.lava_size[0] / 2 + self.platform_size[0] / 2, self.lava_position[1], self.lava_position[2]]
        self.platform_start_id = self.create_box(mass=0., half_extents=[self.platform_size[0] / 2, self.platform_size[1] / 2, self.platform_size[2] / 2], position=self.platform_start_position, color=[0.3, 0.3, 0.3, 1.])
        self.platform_end_id = self.create_box(mass=0., half_extents=[self.platform_size[0] / 2, self.platform_size[1] / 2, self.platform_size[2] / 2], position=self.platform_end_position, color=[0.3, 0.3, 0.3, 1.])
        self._p.changeDynamics(bodyUniqueId=self.platform_start_id, linkIndex=-1, lateralFriction=0.8, restitution=0.5)
        self._p.changeDynamics(bodyUniqueId=self.platform_end_id, linkIndex=-1, lateralFriction=0.8, restitution=0.5)

        # Init boat
        self.boat_size = [3., 2., 0.2]
        self.boat_position_init = [self.lava_position[0] - self.lava_size[0] / 2 + self.boat_size[0] / 2, self.lava_position[1], self.boat_size[2] / 2]
        self.boat_speed = 3.
        self.boat_id = self.create_box(mass=0., half_extents=[self.boat_size[0] / 2, self.boat_size[1] / 2, self.boat_size[2] / 2], position=self.boat_position_init, color=[0.8, 0.8, 0.8, 1.])
        self._p.changeDynamics(bodyUniqueId=self.boat_id, linkIndex=-1, lateralFriction=0.8, restitution=0.5)

        # Init button
        self.button_radius = 0.25
        self.button_height = 0.25
        self.button_position_init = [self.boat_position_init[0] + self.boat_size[0] / 4, self.lava_position[1], self.boat_position_init[2] + self.boat_size[2] / 2 + self.button_height / 2]  # put button on the right side of the boat
        self.button_id = self.create_cylinder(mass=0., radius=self.button_radius, height=self.button_height, position=self.button_position_init, color=[0., 0.5, 0., 1.])

        # Init target zone
        self.target_zone_radius = 1.5
        self.target_zone_id = self.create_sphere(mass=0., radius=self.target_zone_radius, collision=False, position=[self.platform_end_position[0], self.platform_end_position[1], self.platform_end_position[2] + self.platform_size[2] / 2], color=[0., 1., 0., 0.5])

        self.objects_on_boat = [self.button_id]

    def create_box(self, mass, half_extents, position, color):
        collision_shape_id = self._p.createCollisionShape(shapeType=self._p.GEOM_BOX, halfExtents=half_extents)
        visual_shape_id = self._p.createVisualShape(shapeType=self._p.GEOM_BOX, halfExtents=half_extents, rgbaColor=color)
        return self._p.createMultiBody(baseMass=mass, baseCollisionShapeIndex=collision_shape_id, baseVisualShapeIndex=visual_shape_id, basePosition=position)

    def create_cylinder(self, mass, radius, height, position, color):
        collision_shape_id = self._p.createCollisionShape(shapeType=self._p.GEOM_CYLINDER, radius=radius, height=height)
        visual_shape_id = self._p.createVisualShape(shapeType=self._p.GEOM_CYLINDER, radius=radius, length=height, rgbaColor=color)
        return self._p.createMultiBody(baseMass=mass, baseCollisionShapeIndex=collision_shape_id, baseVisualShapeIndex=visual_shape_id, basePosition=position)

    def create_sphere(self, mass, radius, collision, position, color):
        if collision:
            collision_shape_id = self._p.createCollisionShape(shapeType=self._p.GEOM_SPHERE, radius=radius)
            visual_shape_id = self._p.createVisualShape(shapeType=self._p.GEOM_SPHERE, radius=radius, rgbaColor=color)
            return self._p.createMultiBody(baseMass=mass, baseCollisionShapeIndex=collision_shape_id, baseVisualShapeIndex=visual_shape_id, basePosition=position)
        else:
            visual_shape_id = self._p.createVisualShape(shapeType=self._p.GEOM_SPHERE, radius=radius, rgbaColor=color)
            return self._p.createMultiBody(baseMass=mass, baseVisualShapeIndex=visual_shape_id, basePosition=position)

    def create_heightfield(self, size, height_max, position, resolution, repeats=2):
        heightfield_data = np.random.uniform(low=0., high=height_max, size=(int(size[0] * resolution / repeats), int(size[1] * resolution / repeats)))
        heightfield_data = np.repeat(np.repeat(heightfield_data, repeats, axis=0), repeats, axis=1)
        mesh_scale = [1/resolution, 1/resolution, 1.]
        heightfield_collision_shape_id = self._p.createCollisionShape(
            shapeType=self._p.GEOM_HEIGHTFIELD,
            meshScale=mesh_scale,
            heightfieldData=heightfield_data.reshape(-1),
            numHeightfieldRows=heightfield_data.shape[0],
            numHeightfieldColumns=heightfield_data.shape[1],
        )
        return self._p.createMultiBody(baseMass=0., baseCollisionShapeIndex=heightfield_collision_shape_id, basePosition=[position[0], position[1], position[2] + mesh_scale[2] * height_max / 2])

    def get_object_position(self, object_id):
        return np.asarray(self._p.getBasePositionAndOrientation(object_id)[0])

    def get_distance_to_object(self, object_id):
        object_position = self.get_object_position(object_id)
        robot_position = self.robot.links["base"].position
        return np.linalg.norm(object_position[:2] - robot_position[:2])

    def reset(self):
        observation = super().reset()

        # Reset boat position
        boat_y_init = np.random.uniform(low=-self.lava_size[1] / 2 + self.boat_size[1] / 2, high=self.lava_size[1] / 2 - self.boat_size[1] / 2)  # randomize y position
        self._p.resetBasePositionAndOrientation(self.boat_id, [self.boat_position_init[0], boat_y_init, self.boat_position_init[2]], [0., 0., 0., 1.])

        # Reset button position
        self._p.resetBasePositionAndOrientation(self.button_id, [self.button_position_init[0], boat_y_init, self.button_position_init[2]], [0., 0., 0., 1.])

        # Reset target zone
        target_zone_y = np.random.uniform(low=-self.lava_size[1] / 2 + self.target_zone_radius, high=self.lava_size[1] / 2 - self.target_zone_radius)  # randomize y position
        self.target_zone_position = [self.platform_end_position[0], target_zone_y, self.platform_end_position[2] + self.platform_size[2] / 2]
        self._p.resetBasePositionAndOrientation(self.target_zone_id, self.target_zone_position, [0., 0., 0., 1.])

        # Reset robot position
        self._p.resetBasePositionAndOrientation(self.robot.robot_id, [self.platform_start_position[0], self.platform_start_position[1], self.platform_start_position[2] + self.platform_size[2] / 2 + self.robot.links["base"].position_init[2]], self.robot.links["base"].orientation_init)

        return observation

    def step(self, action):
        # Before taking action
        self.distance_to_button = self.get_distance_to_object(self.button_id)
        self.distance_to_target_zone = self.get_distance_to_object(self.target_zone_id)
        self.has_touched_platform_end = len(self._p.getContactPoints(bodyA=self.robot.robot_id, bodyB=self.platform_end_id)) > 0

        observation, reward, terminated, truncated, info = super().step(action)

        # Check if button is pressed
        contact_points = self._p.getContactPoints(bodyA=self.robot.robot_id, bodyB=self.button_id)
        button_pressed = len(contact_points) > 0

        if button_pressed:
            # Move boat and everything on boat forward
            for body_id in [self.boat_id] + self.objects_on_boat:
                body_position = self.get_object_position(body_id)
                new_object_position = body_position + np.array([self.boat_speed * self.dt, 0., 0.])
                self._p.resetBasePositionAndOrientation(body_id, new_object_position, [0., 0., 0., 1.])

        return observation, reward, terminated, truncated, info

    def get_task_rewards(self, action):
        # After taking action
        new_distance_to_button = self.get_distance_to_object(self.button_id)
        new_distance_to_target_zone = self.get_distance_to_object(self.target_zone_id)

        # Survival
        survival = 1.

        # Reach button
        reach_button = (self.distance_to_button - new_distance_to_button) / self.dt

        # Reach target zone
        reach_target_zone = (self.distance_to_target_zone - new_distance_to_target_zone) / self.dt
        if self.distance_to_target_zone < self.target_zone_radius:
            reach_target_zone += 5.

        return {"survival": survival, "reach_button": reach_button, "reach_target_zone": reach_target_zone}

    def get_terminated(self, action):
        # Terminate if touch lava
        contact_points = self._p.getContactPoints(bodyA=self.robot.robot_id, bodyB=self.lava_id)
        is_touching_lava = len(contact_points) > 0

        # Terminate if fall off
        is_fall_off = self.robot.links["base"].position[2] < self.platform_start_position[2]
        return is_touching_lava or is_fall_off

    def get_success(self):
        # Success if stand in the target zone
        distance_to_target_zone = self.get_distance_to_object(self.target_zone_id)
        return distance_to_target_zone < self.target_zone_radius
\end{lstlisting}

\begin{lstlisting}[language=Python]
import numpy as np
from oped.envs.r2d2.base import R2D2Env


class Env(R2D2Env):
    """
    Descend a series of stairs to reach the ground.

    Description:
    - The environment consists of a ground platform (1000 m x 10 m x 10 m) and a set of 10 steps.
    - Each step has dimensions of 1 m in length, 10 m in width, and 0.2 m in height.
    - The steps are positioned to form a descending staircase starting from an initial height, with each subsequent step lower than the previous one.
    The robot is initialized at the top of the stairs.

    Success:
    The task is completed when the robot successfully descends the stairs and touches the ground platform.

    Rewards:
    The help the robot complete the task:
    - The robot is rewarded for survival at each time step.
    - The robot is rewarded for forward velocity, incentivizing it to move down the stairs.

    Termination:
    The task terminates immediately if the robot falls off the stairs or the ground platform.
    """

    def __init__(self):
        super().__init__()

        # Init ground
        self.ground_size = [1000., 10., 10.]
        self.ground_position = [0., 0., 0.]
        self.ground_id = self.create_box(mass=0., half_extents=[self.ground_size[0] / 2, self.ground_size[1] / 2, self.ground_size[2] / 2], position=self.ground_position, color=[0.5, 0.5, 0.5, 1.])
        self._p.changeDynamics(bodyUniqueId=self.ground_id, linkIndex=-1, lateralFriction=0.8, restitution=0.5)

        # Init stairs
        self.num_steps = 10
        self.step_size = [1.0, 10., 0.2]
        self.step_position_init = [self.ground_position[0], self.ground_position[1], self.ground_position[2] + self.ground_size[2] / 2 + self.num_steps * self.step_size[2]]
        self.create_stairs_down(step_size=self.step_size, step_position_init=self.step_position_init, num_steps=self.num_steps)

    def create_box(self, mass, half_extents, position, color):
        collision_shape_id = self._p.createCollisionShape(shapeType=self._p.GEOM_BOX, halfExtents=half_extents)
        visual_shape_id = self._p.createVisualShape(shapeType=self._p.GEOM_BOX, halfExtents=half_extents, rgbaColor=color)
        return self._p.createMultiBody(baseMass=mass, baseCollisionShapeIndex=collision_shape_id, baseVisualShapeIndex=visual_shape_id, basePosition=position)

    def create_stairs_down(self, step_size, step_position_init, num_steps):
        color_1 = np.array([1., 0., 0.])
        color_2 = np.array([0., 0., 1.])
        for i in range(num_steps):
            step_position = [step_position_init[0] + i * step_size[0], step_position_init[1], step_position_init[2] - i * step_size[2]]
            interpolation = i / (num_steps - 1)
            step_color = (1 - interpolation) * color_1 + interpolation * color_2  # shade steps for visualization
            self.create_box(mass=0., half_extents=[step_size[0] / 2, step_size[1] / 2, step_size[2] / 2], position=step_position, color=np.append(step_color, 1.))

    def reset(self):
        observation = super().reset()

        # Reset robot position at the top of the stairs
        self._p.resetBasePositionAndOrientation(self.robot.robot_id, [self.step_position_init[0], self.step_position_init[1], self.step_position_init[2] + self.step_size[2] / 2 + self.robot.links["base"].position_init[2]], self.robot.links["base"].orientation_init)

        return observation

    def step(self, action):
        # Before taking action
        self.position = self.robot.links["base"].position

        observation, reward, terminated, truncated, info = super().step(action)

        return observation, reward, terminated, truncated, info

    def get_task_rewards(self, action):
        # After taking action
        new_position = self.robot.links["base"].position

        # Survival
        survival = 1.

        # Forward velocity
        forward_velocity = (new_position[0] - self.position[0]) / self.dt

        return {"survival": survival, "forward_velocity": forward_velocity}

    def get_terminated(self, action):
        # Terminate if fall off
        return self.robot.links["base"].position[2] < self.ground_position[2]

    def get_success(self):
        # Success if reach end stairs and touch ground
        contact_points = self._p.getContactPoints(bodyA=self.robot.robot_id, bodyB=self.ground_id)
        is_on_ground = len(contact_points) > 0
        return is_on_ground
\end{lstlisting}

\begin{lstlisting}[language=Python]
import numpy as np
from oped.envs.r2d2.base import R2D2Env


class Env(R2D2Env):
    """
    Activate a lever to open a door and move through the door.

    Description:
    - The environment consists of a large platform measuring 1000 x 10 x 0.1 meters.
    - The robot is initialized at a fixed position on the platform.
    - A door with dimensions 0.5 x 2 x 2 meters is positioned on the platform, 5 m aways from the robot, initially closed.
    - The door is flanked by walls to prevent the robot from bypassing it.
    - A lever is placed on the platform, 2 meters to the left of the door.
    - The task of the robot is to move to the lever, activate it to open the door, and then pass through the door.

    Success:
    The task is successfully completed if the robot passes through the door and moves more than 10 m beyond the initial position.

    Rewards:
    To guide the robot to complete the task:
    - The robot receives a survival reward at each time step.
    - The robot is rewarded for decreasing its distance to the lever.
    - The robot receives a bonus rewards for activating the lever to open the door.
    - Once the door is open, the robot is rewarded for moving forward.

    Termination:
    The task terminates immediately if the robot falls off the stairs or the ground platform.
    """

    def __init__(self):
        super().__init__()

        self.robot_position_init = [0., 0., 0.]

        # Init platform
        self.platform_size = [1000., 10., 0.1]
        self.platform_position = [self.robot_position_init[0] + self.platform_size[0] / 2 - 2., self.robot_position_init[1], self.robot_position_init[2] - self.platform_size[2] / 2]  # offset by 2 m to avoid off-edge or on-edge placement
        self.platform_id = self.create_box(mass=0., half_extents=[self.platform_size[0] / 2, self.platform_size[1] / 2, self.platform_size[2] / 2], position=self.platform_position, color=[0.5, 0.5, 0.5, 1.])
        self._p.changeDynamics(bodyUniqueId=self.platform_id, linkIndex=-1, lateralFriction=0.8, restitution=0.5)

        # Init door
        self.door_size = [0.5, 2., 2.]
        self.door_position_init = [self.robot_position_init[0] + 5., self.platform_position[1], self.platform_position[2] + self.platform_size[2] / 2 + self.door_size[2] / 2]
        self.door_id = self.create_box(mass=0., half_extents=[self.door_size[0] / 2, self.door_size[1] / 2, self.door_size[2] / 2], position=self.door_position_init, color=[1., 0., 0., 1.])
        self.door_open = False

        # Init wall
        self.wall_size = [self.door_size[0], (self.platform_size[1] - self.door_size[1]) / 2, self.door_size[2]]  # walls plus door span the full platform to prevent robot to go around
        self.create_box(mass=0., half_extents=[self.wall_size[0] / 2, self.wall_size[1] / 2, self.wall_size[2] / 2], position=[self.door_position_init[0], self.door_position_init[1] + self.door_size[1] / 2 + self.wall_size[1] / 2, self.platform_position[2] + self.platform_size[2] / 2 + self.wall_size[2] / 2], color=[0., 0., 1., 1.])  # left section
        self.create_box(mass=0., half_extents=[self.wall_size[0] / 2, self.wall_size[1] / 2, self.wall_size[2] / 2], position=[self.door_position_init[0], self.door_position_init[1] - self.door_size[1] / 2 - self.wall_size[1] / 2, self.platform_position[2] + self.platform_size[2] / 2 + self.wall_size[2] / 2], color=[0., 0., 1., 1.])  # right section

        # Init lever
        self.lever_radius = 0.05
        self.lever_height = 0.5
        lever_position = [self.door_position_init[0] - 2., self.door_size[1], self.platform_position[2] + self.platform_size[2] / 2 + self.lever_height / 2]  # two meters to the left of the door on the platform
        self.lever_id = self.create_cylinder(mass=0., radius=self.lever_radius, height=self.lever_height, position=lever_position, color=[0.5, 0.25, 0., 1.])

    def create_box(self, mass, half_extents, position, color):
        collision_shape_id = self._p.createCollisionShape(shapeType=self._p.GEOM_BOX, halfExtents=half_extents)
        visual_shape_id = self._p.createVisualShape(shapeType=self._p.GEOM_BOX, halfExtents=half_extents, rgbaColor=color)
        return self._p.createMultiBody(baseMass=mass, baseCollisionShapeIndex=collision_shape_id, baseVisualShapeIndex=visual_shape_id, basePosition=position)

    def create_cylinder(self, mass, radius, height, position, color):
        collision_shape_id = self._p.createCollisionShape(shapeType=self._p.GEOM_CYLINDER, radius=radius, height=height)
        visual_shape_id = self._p.createVisualShape(shapeType=self._p.GEOM_CYLINDER, radius=radius, length=height, rgbaColor=color)
        return self._p.createMultiBody(baseMass=mass, baseCollisionShapeIndex=collision_shape_id, baseVisualShapeIndex=visual_shape_id, basePosition=position)

    def get_object_position(self, object_id):
        return np.asarray(self._p.getBasePositionAndOrientation(object_id)[0])

    def get_distance_to_object(self, object_id):
        object_position = self.get_object_position(object_id)
        robot_position = self.robot.links["base"].position
        return np.linalg.norm(object_position[:2] - robot_position[:2])

    def reset(self):
        observation = super().reset()

        # Reset door
        self.door_open = False
        self._p.resetBasePositionAndOrientation(self.door_id, self.door_position_init, [0., 0., 0., 1.])

        # Reset robot position
        self._p.resetBasePositionAndOrientation(self.robot.robot_id, [self.robot_position_init[0], self.robot_position_init[1], self.robot_position_init[2] + self.robot.links["base"].position_init[2]], self.robot.links["base"].orientation_init)

        return observation

    def step(self, action):
        # Before taking action
        self.position = self.robot.links["base"].position
        self.distance_to_lever = self.get_distance_to_object(self.lever_id)

        observation, reward, terminated, truncated, info = super().step(action)

        contact_points = self._p.getContactPoints(bodyA=self.robot.robot_id, bodyB=self.lever_id)
        if len(contact_points) > 0 and not self.door_open:
            self.door_open = True
            self._p.resetBasePositionAndOrientation(self.door_id, [self.door_position_init[0], self.door_position_init[1] + self.door_size[1], self.door_position_init[2]], [0., 0., 0., 1.])

        return observation, reward, terminated, truncated, info

    def get_task_rewards(self, action):
        # After taking action
        new_position = self.robot.links["base"].position
        new_distance_to_lever = self.get_distance_to_object(self.lever_id)

        # Survival
        survival = 1.

        # Reach lever
        if not self.door_open and len(self._p.getContactPoints(bodyA=self.robot.robot_id, bodyB=self.lever_id)) == 0:
            reach_lever = (self.distance_to_lever - new_distance_to_lever) / self.dt
        elif not self.door_open and len(self._p.getContactPoints(bodyA=self.robot.robot_id, bodyB=self.lever_id)) > 0:
            reach_lever = 10.
        else:
            reach_lever = 0.

        # Forward velocity
        if self.door_open:
            forward_velocity = (new_position[0] - self.position[0]) / self.dt
        else:
            forward_velocity = 0.

        return {"survival": survival, "reach_lever": reach_lever, "forward_velocity": forward_velocity}

    def get_terminated(self, action):
        # Terminate if fall off
        return self.robot.links["base"].position[2] < self.platform_position[2]

    def get_success(self):
        # Success if pass through door
        return self.robot.links["base"].position[0] > 10.
\end{lstlisting}

\newpage
\section{Task Description Seeds}
\label{appendix:task-desc-seeds}
\begin{lstlisting}[language={}]
Cross a pride-colored bridge with gaps to reach a platform.

Description:
- A start platform and an end platform (each 3 m in size and 0.5 m in thickness) are placed 50 m apart.
- The two platforms are connected by a bridge (2 m wide) divided in multiple segments. Each segment has a different color corresponding to the pride colors.
- The segments are separated by gaps measuring 2 m.
The robot is initialized on the start platform.
The task of the robot is to cross the bridge to reach the end platform as fast as possible.

Success:
The task is successfully completed when the robot reaches the end platform.

Rewards:
To help the robot complete the task:
- The robot receives a reward for each time step it remains standing on the bridge or platforms, encouraging steady progress.
- The robot is rewarded based on how much it reduces the distance to the end platform, incentivizing swift movement towards the goal.

Termination:
The task terminates immediately if the robot falls off the start platform, any segment of the bridge, or the end platform.
\end{lstlisting}

\begin{lstlisting}[language={}]
Ascend a series of stairs to reach a platform.

Description:
- The environment consists of a ground platform (1000 m x 10 m x 10 m) and a set of 10 steps.
- Each step has dimensions of 1 m in length, 10 m in width, and 0.2 m in height.
- The steps are positioned to form an ascending staircase, with each subsequent step higher than the previous one.
The robot is initialized on the ground at the bottom of the stairs.

Success:
The task is completed when the robot successfully ascends the stairs and reaches the top platform.

Rewards:
To help the robot complete the task:
- The robot is rewarded for survival at each time step.
- The robot is rewarded for forward velocity, incentivizing it to move up the stairs.

Termination:
The task terminates immediately if the robot falls off the stairs or the top platform.
\end{lstlisting}

\begin{lstlisting}[language={}]
Kick a ball into a goal.

Description:
- The environment consists of a large flat ground measuring 1000 x 1000 x 10 meters.
- A ball with a radius of 0.5 meters is placed randomly on the ground.
- The goal is defined by two goal posts, each 2 meters high and placed 3 meters apart, forming a goal area.
- The robot is initialized at a fixed position on the ground.
- The task of the robot is to move across the ground, reach the ball, and kick it into the goal.

Success:
The task is successfully completed if the robot kicks the ball so that it passes between the two goal posts.

Rewards:
To help the robot complete the task:
- The robot is rewarded for survival at each time step.
- The robot is rewarded for decreasing its distance to the ball.
- The robot is rewarded for kicking the ball towards the goal, with additional rewards for successfully kicking the ball into the goal.

Termination:
The task does not have a specific termination condition.
\end{lstlisting}

\newpage
\section{Discussion on Training Generalists}
\label{appendix:generalists}
Previous work~\citep{bauer_HumanTimescaleAdaptationOpenEnded_2023} has shown that training a large model on a sufficiently diverse set of tasks can produce a generalist agent capable of generalizing to new tasks within that distribution. However, generating such a distribution is both time-consuming and challenging. Furthermore, to achieve efficiency, it is crucial to automatically create a curriculum within this distribution. Since these are the main challenges, our results section focuses on demonstrating that \ours{} addresses both issues effectively (\Cref{sec:short-run}).
While our academic lab lacks the resources to train generalist agents on a large set of \ours{}-generated tasks, our findings suggest that it should be feasible to do so, as \ours{} generates a progressively complex and diverse curriculum. Nonetheless, we acknowledge that training such agents would still be a significant endeavor, as demonstrated by \citet{bauer_HumanTimescaleAdaptationOpenEnded_2023}, which required an extensive distribution of environments, considerable effort, and substantial computational resources.

\newpage
\section{Experiments on Different Robots and Action Spaces}
\label{appendix:results-ant}

OMNI-EPIC is designed to accommodate a wide range of robotic systems, regardless of the robot type or action space. To demonstrate the flexibility of our approach, we train an Ant robot on generated tasks using OMNI-EPIC. The Ant robot is a 3D quadruped consisting of a torso with free rotational movement and four legs, each composed of two segments \citep{towers_gymnasium_2023}. Apart from the robot type and action space, all settings, including the observation space and RL algorithm, are kept consistent with those used for the R2D2 robot. The Ant robot's action space is defined as a continuous 8-dimensional vector, with each element bounded between -1 and 1.

\begin{figure}[ht]
    \centering
    \includegraphics[width=\textwidth]{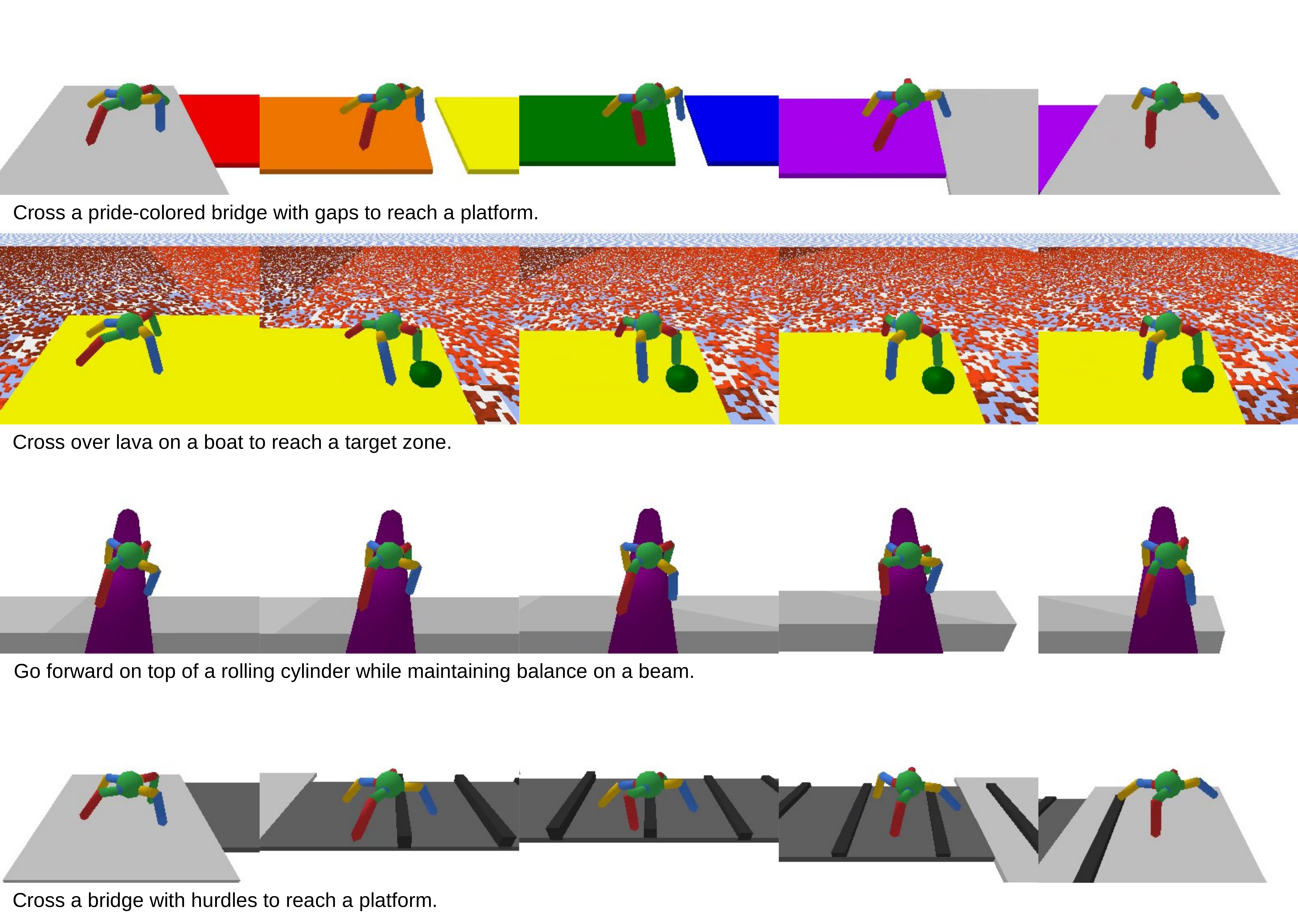}
    \caption{\textbf{Ant robot successfully completing different generated tasks.} These examples highlight OMNI-EPIC's ability to train various robot types and operate across different action spaces.}
    \label{fig:results-ant}
\end{figure}

\end{document}